%% file: main.tex
\pdfoutput=1

\documentclass[11pt]{article}

\usepackage[]{acl}

\usepackage{times}
\usepackage{latexsym}

\usepackage{graphicx}
\usepackage{amsmath}
\usepackage{booktabs}
\usepackage{cleveref}
\usepackage{multirow}
\usepackage{hyperref}
\usepackage{amssymb}
\usepackage{pifont}
\usepackage{color, colortbl}
\usepackage{comment}
\usepackage{subcaption}
\usepackage{amsmath}
\RequirePackage{algorithm}
\RequirePackage{algorithmic}

\usepackage[T1]{fontenc}

\usepackage[utf8]{inputenc}

\usepackage{microtype}
\usepackage{bm}

\usepackage[utf8]{inputenc}

\input{./math_commands.tex}

\definecolor{Color}{gray}{0.96}
\definecolor{Color1}{gray}{0.89}

\title{KDMCSE: Knowledge Distillation Multimodal Sentence Embeddings with Adaptive Angular margin Contrastive Learning} 

\author{Cong-Duy Nguyen$^{1}$,~~Thong Nguyen$^{2}$,~~Xiaobao Wu$^{1}$,~~Luu Anh Tuan$^{1}$\\
  $^1$Nanyang Technological University, Singapore \\
  $^2$National University of Singapore, Singapore}
    
\begin{document}
\maketitle

\input{sections/abstract}
\input{sections/introduction}
\input{sections/related_work}
\input{sections/method}
\input{sections/experiments}

\input{sections/analysis}

\input{sections/conclusion}
\input{sections/limitations}

\section*{Acknowledgement}
We thank all anonymous reviewers for their helpful comments. This research/project is supported by the National Research Foundation, Singapore under its AI Singapore Programme, AISG Award No: AISG2-TC-2022-005.

\bibliography{anthology}
\bibliographystyle{acl_natbib}
\appendix
\input{sections/appendix}

\clearpage

\end{document}

%% file: math_commands.tex
\def\eqref#1{equation~\ref{#1}}

\def\1{\bm{1}}

\def\vh{{\bm{h}}}

\def\vm{{\bm{m}}}

\def\vs{{\bm{s}}}
\def\vt{{\bm{t}}}

\def\vv{{\bm{v}}}

\DeclareMathAlphabet{\mathsfit}{\encodingdefault}{\sfdefault}{m}{sl}
\SetMathAlphabet{\mathsfit}{bold}{\encodingdefault}{\sfdefault}{bx}{n}

%% file: sections/abstract.tex
\begin{abstract}

Previous work on multimodal sentence embedding has proposed multimodal contrastive learning and achieved promising results. However, by taking the rest of the batch as negative samples without reviewing when forming contrastive pairs, those studies encountered many suspicious and noisy negative examples, significantly affecting the methods' overall performance. In this work, we propose KDMCSE (Knowledge Distillation Multimodal contrastive learning of Sentence Embeddings), a novel approach that enhances the discrimination and generalizability of multimodal representation and inherits the knowledge from the teacher model to learn the difference between positive and negative instances and via that, can detect noisy and wrong negative samples effectively before they are calculated in the contrastive objective. Furthermore, to overcome the limitation of modeling the variation within negative pairs, we introduce a new contrastive objective, AdapACSE (Adaptive Angular Margin Supervised Contrastive Learning for Multimodal sentence embeddings), that enhances the discriminative representation by strengthening the margin within the angular space while capturing varying semantics within the negative. Experimental results on widely used Semantic Textual Similarity (STS) benchmarks demonstrate the effectiveness of our approach. The source code is available at \url{https://github.com/duyngtr16061999/KDMCSE}.

\end{abstract}

%% file: sections/introduction.tex
\section{Introduction}

Learning sentence embeddings, which involves translating sentences into consistent-length vectors that capture their semantic connections, remains a pivotal task in the NLP field. While pre-trained language models like BERT~\citep{devlin2019bert} and RoBERTa~\citep{liu2019roberta} have witnessed immense success, studies indicate that the default sentence embeddings from these PLMs, without specific fine-tuning, may not perform as well as simply averaging Glove vectors~\citep{pennington2014glove} in capturing semantic similarity~\citep{reimers-gurevych-2019-sentence}. Consequently, recent works~\citep{li-etal-2020-sentence, zhang-etal-2020-unsupervised, su2021whitening} are geared towards refining sentence embeddings from PLMs without supervision.

\begin{figure}[t]
    \centering
    \includegraphics[width=0.48\textwidth]{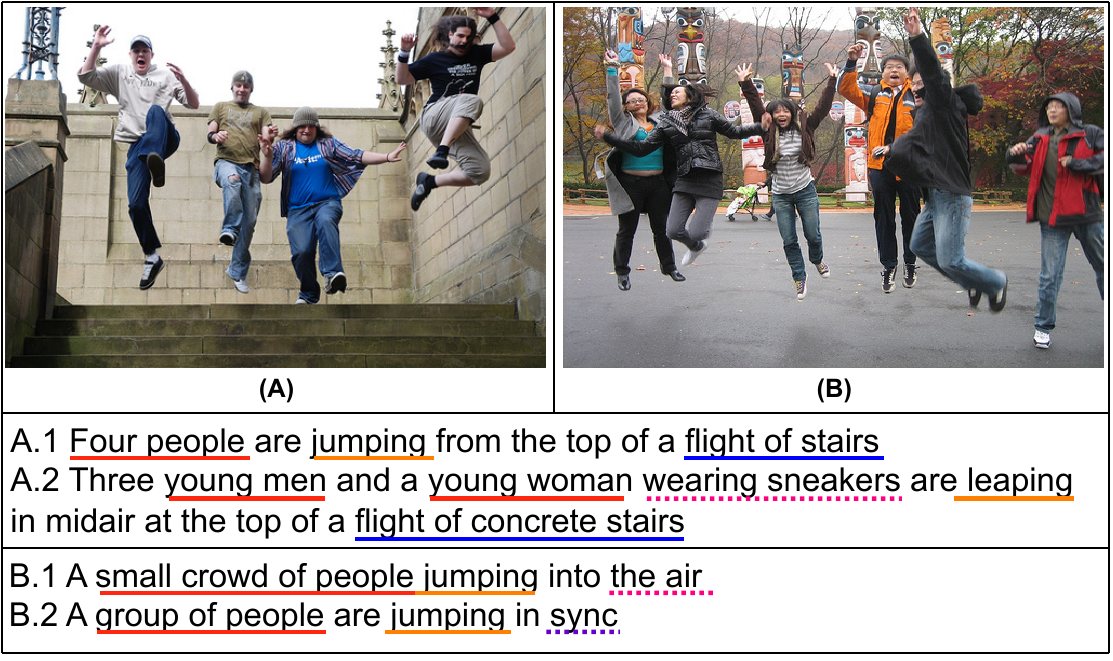} 
    \caption{Example image-caption pairs in Flickr. Solid lines of the same color talk about the same instance, and a dot line means the additional information that does not occur in the other caption. }
    \vspace{-6pt}
    \label{fig:vis1}
\end{figure}

\begin{figure}[t]
    \centering
    \includegraphics[width=0.4\textwidth]{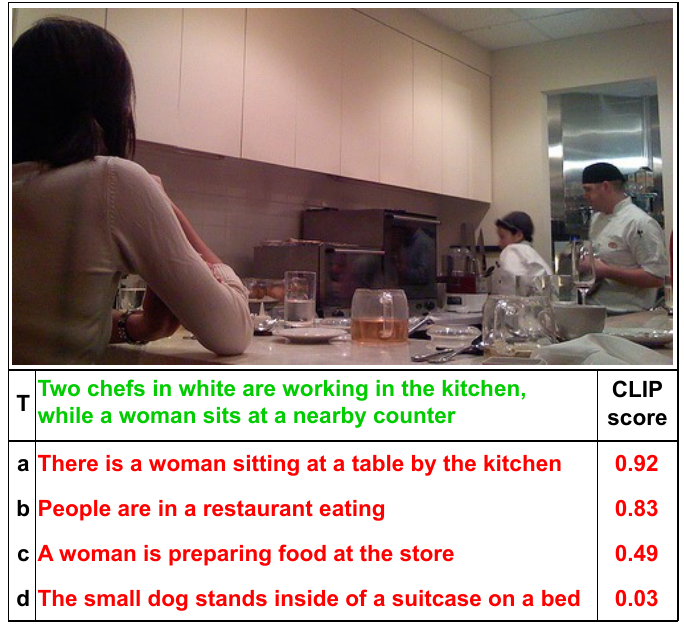} 
    \caption{Example image-caption pairs in Flickr. The green caption is a true annotation of the image while 4 red captions is randomly picked from the dataset. The scores on the right are the cosine similarity between its caption representation and image representation extracted from CLIP model.}
    \label{fig:vis2}
    \vspace{-6pt}
\end{figure}


While text-centric models have made significant strides, their depth of understanding of sentence semantics remains a point of contention. True semantic understanding often stems from associations in the real world rather than mere textual statistics~\citep{wang2019holistic,bender-koller-2020-climbing, bisk-etal-2020-experience,wei2022audio}.
Numerous recent studies have explored the enhancement of language representations through visual information~\cite{vokenization, tang2021vidlankd,nguyen2024expand}. 
The recent multimodal contrast in sentence embedding learning~\cite{zhang2022mcse} offers an encouraging direction.
This method augments SimCSE with a multimodal contrastive objective, 
which seeks congruence between sentences and their associated images within a unified space. However, samples drawn randomly from grounded datasets during training can present semantic differences, subtle similarities, or overlapping attributes. For example, in Figure~\ref{fig:vis1}, two captions B.1 and B.2 describe the second image B, but also correctly represent the first image A. Due to their inherent noise, such samples can hinder the learning process. Mitigating this issue requires preemptively filtering out these semantically similar samples before loss calculation, ensuring a noise-reduced training environment. In addition, comparing two captions, A.1 and A.2, for the left image reveals that the latter offers a more comprehensive description, detailing aspects such as the individual's gender, attire, sneakers, and even the material of the stairs (concrete). Thus, using only visual images cannot capture the depth of language representations.

Furthermore, MCSE did not consider the distinctions between negative sample pairs while focusing on constructing pairs of positive and negative representations.
As shown in Figure~\ref{fig:vis2}, although none of the captions perfectly describes the image, their levels of discrimination vary between pairs. 
On the contrary, the caption $c$ registers a score of $0.49$, hinting at its partial relevance to the image. Caption $d$, with a very weak score of $0.03$, is completely unrelated. Consequently, within the feature space, both captions $c$ and $d$ should be substantially distanced from the accurate depiction. Furthermore, the gap between the true caption and caption d should considerably exceed that between the true caption and the caption $c$.


To address the aforementioned limitation of noisy negative sampling and complexity of linguistic representation, we present KDMCSE: a Knowledge Distillation Multimodal Contrastive Learning framework for Sentence Embeddings that utilizes CLIP~\cite{radford2021clip} as the teacher model for both image and text modalities. Our approach harnesses images for multimodal objectives and textual representations derived from the CLIP text encoder. Relying solely on images falls short of capturing the nuances in sentence representation. Emulating the representation of the CLIP's textual encoder - especially when paired with image matching - can augment the richness of linguistic representations, transcending a mere alignment with visual features.

Moreover, in response to the identified challenge of diversity among negative pairs, we introduce AdapACSE: Adaptive Angular Margin Contrastive Learning. This method bolsters discriminative representation by amplifying the margin within the angular space, specifically accounting for the various semantics present in negative samples. Leveraging the teacher model, CLIP, we produce soft labels that signify the similarity between samples. Instead of merely utilizing feature representations as positive or negative samples during the multimodal objective, we employ CLIP's soft labels to intensely penalize samples exhibiting low similarity and provide leniency to pairs exhibiting certain resemblances. 

Our contributions can be summarized as follows.

\begin{itemize}
\item We propose a novel multimodal framework for sentence embedding KDMCSE: Knowledge Distillation Multimodal Contrastive learning for Sentence Embedding. The knowledge of the teacher model is transferred to the student model through multimodal contrastive learning, spanning both text and visual modalities.

\item We introduce self-supervised contrastive learning: AdapACSE that improves the discriminative representation of samples with varying degrees of similarity.

\item We evaluate our approach using standard semantic textual similarity (STS) benchmarks and SentEval transfer tasks. Our method outperforms the performance of earlier state-of-the-art approaches.
\end{itemize}

%% file: sections/related_work.tex
\section{Related Work}

\paragraph{Sentence Representation Learning} 

In previous research, sentence representations were typically learned by extending the principles of word2vec~\cite{DBLP:journals/corr/MikolovSCCD13}, predicting neighboring sentences~\cite{DBLP:journals/corr/KirosZSZTUF15,hill2016learning,logeswaran2018efficient} or aggregating n-gram embeddings~\cite{Pagliardini_2018}, the pre-trained transformer-based model BERT~\cite{li-etal-2020-sentence,reimers-gurevych-2019-sentence} and its enhancements BERT-Flow~\cite{li-etal-2020-sentence} and BERT-Whitening~\cite{su2021whitening}. More recently, various studies have embraced the contrastive learning framework for learning sentence representation. These studies have proposed different methods to form contrastive pairs, either through various data transformation techniques~\cite{yan-etal-2021-consert,zhang-etal-2020-unsupervised,giorgi-etal-2021-declutr} or through encoders with distinct structures or parameters~\cite{carlsson2021semantic,kim-etal-2021-self}. A notable example is SimCSE~\cite{gao2021simcse}, which employs dropout as a data augmentation strategy. Many studies in the field of sentence contrastive learning have made a significant contribution by focusing on the mining of challenging negative samples~\cite{Wang_2022,zhou2022debiased, zhang2022unsupervised, wang2022sncse, wei2023multi, he2023instance}.

\paragraph{Deep Metric Learning Objectives}

Contrastive learning \cite{1467314} has gained traction across multiple domains~\cite{nguyen2021contrastive,nguyen2022adaptive,nguyen2024topic,nguyen2023improving,Wu2020short,wu2022mitigating,wu2023infoctm,wu2023effective,wu2024traco,wei2024learning}. Renowned training objectives such as N-Pair Loss \cite{NIPS2016_6b180037}, Triplet Margin Loss \cite{Balntas2016LearningLF}, and ArcCon \cite{zhang-etal-2022-contrastive} are deeply rooted in metric learning principles. In supervised tasks, objectives that leverage softmax have shown efficacy, especially when they integrate class centers and impose penalties on the distances between deep features and their corresponding centers. Among these, Center loss \cite{10.1007/978-3-319-46478-7_31}, SphereFace \cite{8100196}, CosFace \cite{DBLP:journals/corr/abs-1801-09414}, and ArcFace \cite{DBLP:journals/corr/abs-1801-07698} stand out and are frequently used in computer vision and natural language processing applications. Nonetheless, these loss functions primarily aim at classification tasks, rendering them unsuitable for regression labels. 
ArcCSE~\cite{zhang-etal-2022-contrastive} presents training objectives using additional margin tailored to enhance the discriminative prowess in pairwise relations and capture the entailment dynamics within triplet sentence structures. 

\paragraph{Visually Grounded Representation Learning}

Numerous studies have underscored the advantages of integrating NLP models with visual insights to improve textual representation learning (\citet{lazaridou2015combining, kiela2018learning}). Based on the Skip-Thought model of \citet{kiros2015skip}, \citet{bordes2019incorporating} created a unified space that harmoniously accommodates both visual and textual dimensions. Recent pioneering efforts by \citet{tan2020vokenization}, \citet{tang2021vidlankd} and \cite{nguyen2024expand} have laid the foundation for vast language models using multimodal guidance, aiming to improve language understanding. Similarly, MCSE by \citet{zhang2022mcse} suggested sentence embedding learning using a multimodal contrastive objective to align sentences with their respective images coherently. Distinguishing our work from \citet{zhang2022mcse}, we present an innovative angular margin contrastive learning framework, building upon the cutting-edge multimodal contrastive learning proposed by \citet{zhang2022mcse}. By weaving in this new contrastive technique, our goal is to push the envelope in STS by leveraging multimodal semantic data.

%% file: sections/method.tex
\begin{figure*}[t]
    \centering
    \includegraphics[width=0.92\textwidth]{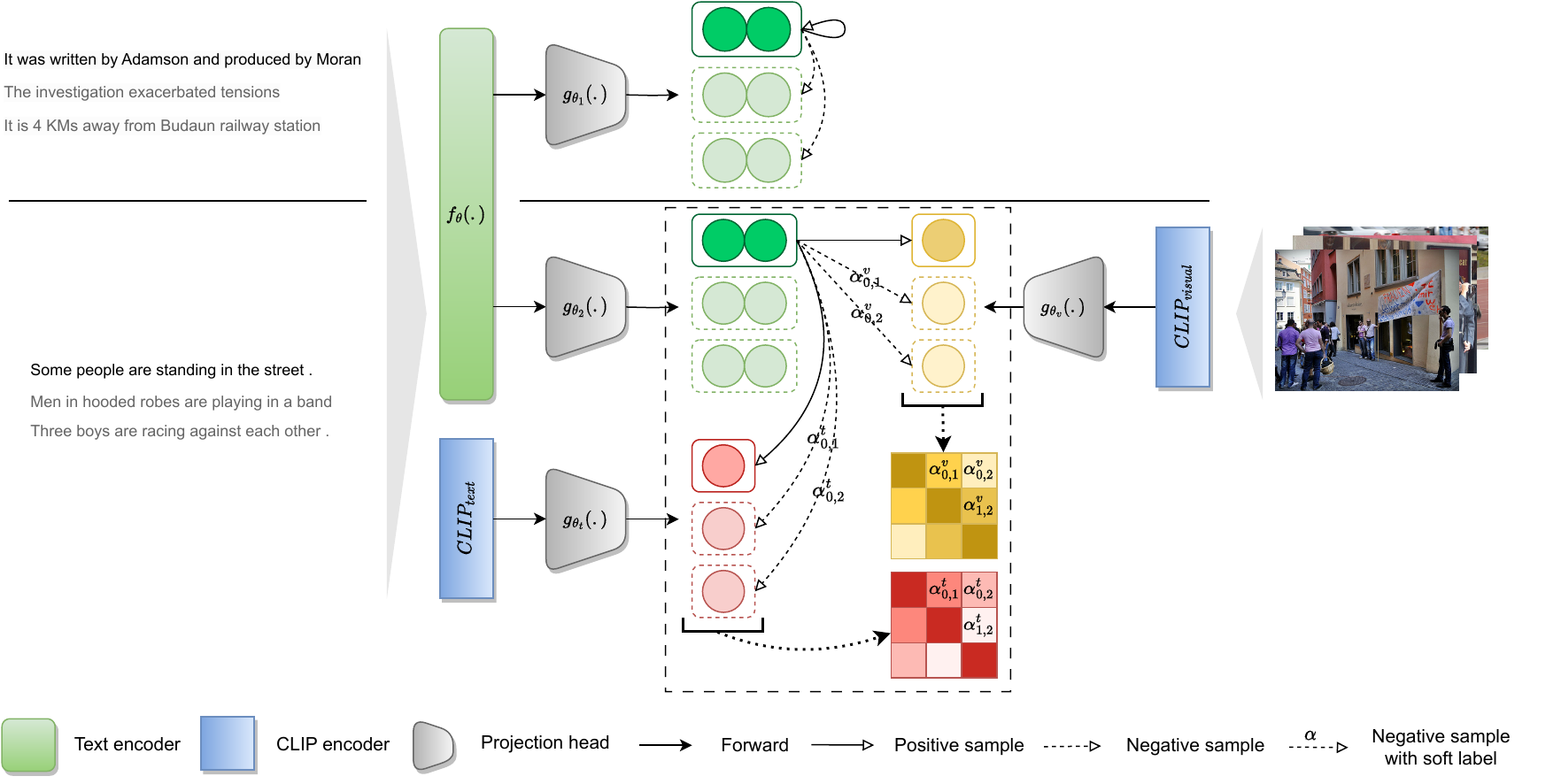} 
    \caption{The overall architecture of KDMCSE. The upper part is the original SimCSE, the below part is the multimodal contrastive learning approach with knowledge distillation from CLIP model. }
    \label{fig:main}
    \vspace{-10pt}
\end{figure*}

\section{Method}
\subsection{Background: Unsupervised \textbf{SimCSE} and Multimodal Contrastive Learning \textbf{MCSE}}
\label{sec:simcse}
The idea of unsupervised SimCSE is to use dropout noise as a simple, yet effective, data augmentation strategy. Given a collection of sentences $\{x_i\}_{i=1}^m$, we construct a positive pair for each input $x_i$ by encoding it twice using different dropout masks: $\vh_i^{z}=g_{\phi}(f_{\theta}(x_i, z))$ and $\vh_i^{z'}=g_{\phi}(f_{\theta}(x_i, z'))$, where $z$ and $z'$ denote different dropout masks\footnote{The standard dropout masks in Transformers are used.}, $f_{\theta}(\cdot)$ is a pre-trained language encoder such as BERT, and $g_{\phi}(\cdot)$ is a projection head\footnote{There is a MLP pooler layer over \texttt{[CLS]} in BERT's implementation. \citet{gao2021simcse} use it with reinitialization.} on top of the \texttt{[CLS]} token. The training objective is: %

\begin{equation}
    \label{eq:cons}
    \ell_i^\mathrm{S}=-\log \frac{e^{\mathrm{sim}(\vh_i^{z_i}, \vh_i^{z_i'})/ \tau}}{ \sum_{j=1}^N e^{\mathrm{sim}(\vh_i^{z_i}, \vh_j^{z_j'})/ \tau}} \;,
\end{equation}

where $N$ is the size of the mini-batch, $\tau$ is a temperature parameter, and $\mathrm{sim} (\vh_1, \vh_2)$ is the cosine similarity $\frac{\vh_1^T \vh_2}{ \left\|\vh_1 \right\|\cdot \left\|\vh_2 \right\|  }$. After training, the \texttt{[CLS]} token outputs of the language encoder are taken as the sentence embeddings.

\label{sec:mcse}
To take advantage of visual and textual information, MCSE~\citep{zhang2022mcse} adopt SimCSE as the textual baseline and extend it with a multimodal contrastive learning objective:
\begin{equation}
    \vs_i^z = g_{\phi_1}(f_{\theta}(x_i, z)),  \,
    \vv_i = g_{\phi_2} (f^v(y_i)) \; ,
\end{equation}
where $f^v(\cdot)$ is a pre-trained image encoder. $g_{\phi_1}(\cdot)$ and $g_{\phi_2}(\cdot)$ are two projection heads for text and image modalities. They define the multimodal contrastive learning objective as:
\begin{equation}
\label{eq:mcse}
    \ell_i^\mathrm{M}=- \sum_{z\in\{ z_i, z_i' \}}\log \frac{e^{\mathrm{sim}(\vs_i^z, \vv_i)/ \tau'}}{ \sum_{j=1}^N e^{\mathrm{sim}(\vs_i^z,\vv_j)/ \tau'}} \; ,
\end{equation}
\subsection{Knowledge Distillation Multimodal Contrastive learning for Sentence Embedding}
\label{sec:kdmcse}  


In addition to the multimodal objective, we present a Knowledge Distillation framework designed to harness both visual and textual insights from the vision-language model CLIP. An overview of our KDMCSE model can be found in Figure \ref{fig:main}. Given a set of sentence-image pairs represented as $D=\{ x_i, y_i  \}_{i=1}^m$, we initially project the sentence $x_i$ and the image $y_i$ into a shared or grounded space.

\begin{equation}
    \vs_i^z = g_{\phi_g}(f_{\theta}(x_i, z)),
\end{equation}
\begin{equation}
    \vt_i = g_{\phi_t}(\hat{\vt_i}),  \,
    \hat{\vt_i} = CLIP_{text}(x_i) \; ,
\end{equation}
\begin{equation}
    \vv_i = g_{\phi_v} (\hat{\vv_i}),  \,
    \hat{\vv_i} = CLIP_{visual}(y_i) \; ,
\end{equation}

\noindent where $CLIP_{text}, CLIP_{visual}$ are a pre-trained text and image CLIP encoder, $\hat{\vv_i}$ and $\hat{\vt_i}$ are the visual and text representation extracted from CLIP, $g_{\phi_g}(\cdot)$ is the projection heads for the language student model to project the sentence representation into grounded space, $g_{\phi_t}(\cdot)$ and $g_{\phi_v}(\cdot)$ are distinct projection heads of the teacher model for the text and image modality, respectively. During training, we freeze the pre-trained CLIP model. 

\begin{figure*}[t]
    \centering
    \includegraphics[width=0.92\textwidth]{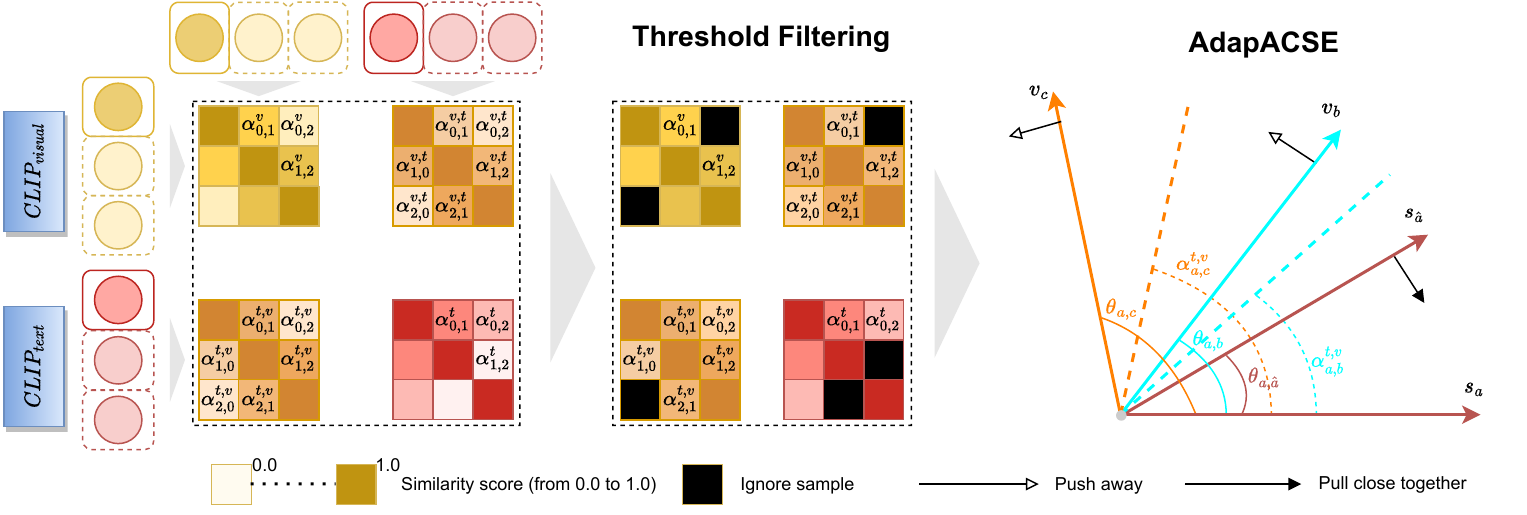} 
    \caption{The overall framework of knowledge distillation with Adaptive Angular margin contrastive learning. 
    The pipeline first calculates the soft-label similarity scores between text and visual representation, then we apply threshold filtering to remove the noisy negative pairs, and finally, we transfer the soft-label matrices into our proposed AdapACSE to flexibly find the margin.
    $s_{\hat{a}}$ is the positive sample for $s_a$, $v_b$ and $v_c$ are its negative counterparts. In particular, the difference in pairs $s_a-v_c$ is more pronounced than in $s_a-v_b$. As a result, the margin (depicted as a dashed line) for $c$ (in orange) is greater than for $b$ (in cyan).}
    \label{fig:kdmcse}
    \vspace{-10pt}
\end{figure*}

\textbf{Threshold Filtering}
During training, negative samples were randomly taken from data in Eq.\ref{eq:mcse}. We not only transfer CLIP’s knowledge of the text and visual representation of each sample, but also leverage the similarity within samples in a batch. We calculate the cosine similarity:

\begin{equation}
\alpha^{m,n}_{i,j} = \dfrac {m_i \cdot n_j} {\left\| m_i\right\| _{2}\left\| n_j\right\| _{2}},  \,
m,n \in \{\hat{v},\hat{t}\}
\end{equation}

\noindent where $m,n$ are the defined modality (text or visual). Thus, as shown in Figure~\ref{fig:kdmcse}, we can generate four similarity mappings: text-text, visual-visual, text-visual, and visual-text. As we strive to align the sentence representation generated by the language encoder (text) with the information gleaned from the teacher model encompassing both sentence (text) and image (visual) data, we exclusively employ 'text-text' and 'text-visual' pairings for our soft labels. We define a threshold filter function to remove noisy negative samples:

\begin{equation}
\varphi^{m,n}_{i,j} = \left\{\begin{matrix}
0 & if & \alpha^{m,n}_{i,j} < threshold \\ 
1 & else &
\end{matrix}\right.
\end{equation}

\noindent where $threshold$ is a fixed hyperparameter during training. The proposed contrastive objective with Threshold Filtering is defined as follows:

\begin{equation}
    \label{eq:phi}
    \ell_i^\mathrm{m}=- \sum_{z\in\{ z_i, z_i' \}}\log \frac{e^{\mathrm{sim}(\vs_i^z, \vm_i)/ \tau'}}{ \sum_{j=1}^N \varphi^{t,m}_{i,j} e^{\mathrm{sim}(\vs_i^z,\vm_j)/ \tau'}} \; ,
\end{equation}

\noindent where $m$ is the modality (text or visual), $m \in \{t,v\}$, $\boldsymbol{m}_{i}$ is the projected representation of CLIP model, $\boldsymbol{m}_{i} \in \{\vv_i, \vt_i\}$.


\subsection{Adaptive Angular margin Contrastive learning}
\label{sec:aabmcl}

Previous studies~\cite{zhang-etal-2022-contrastive} introduced the ArcCSE objective. This was designed to increase the discriminative capacity between pairs and encapsulate the entailment relationship among triplet sentences. ArcCSE adds an additive angular margin $m_c$ between positive pairs $h_i^z$ and $h_i^{z'}$. Compared to Eq.\ref{eq:cons}, it further pushed $h_i$ towards the area where $\theta _{i,i^{'}}$ becomes smaller and $\theta _{j,i}$ becomes larger, increasing the compactness of sentence representations with the same semantics and enlarging the discrepancy of different semantic representations. ArcCSE is defined as:

\begin{align}
\label{eq:arccos}
\ell_i^\mathrm{ArcCSE} = -\mathrm{log} \frac{ e^{\phi\left ( \theta _{i,i^{'}} + m_c \right )/ \tau}}{e^{\phi\left ( \theta _{i,i^{'}} + m \right )/ \tau} + \sum_{j\neq i}^{n} e^{\phi\left ( \theta _{j,i} \right )/ \tau}} 
\end{align}

\noindent where $\phi$ is the $cos$ function, angular $\theta _{i,j}$ is denoted as follows:

\begin{align}
\theta _{i,j} = \mathrm{arccos} \left ( \frac{h_{i}^{T}h_{j}}{\left \| h_{i} \right \| * \left \| h_{j} \right \|} \right )
\end{align}


\noindent where $h_i$ is the vector representation, in this work, $h_i \in \{\vv_i, \vt_i, \vs_i^z\}$.

There are distinct differences among the negative pairs, especially when we observe that the examples marked in red in Figure~\ref{fig:vis2} serve as negative samples for the green caption. In this grounded space, the distance between the samples becomes crucial. We have refined the contrastive objective to harness the knowledge from the teacher model and to capture the magnitude of the differences between negative pairs. To address this challenge, we introduce a novel training objective, Adaptive Angular margin Contrastive Loss, AdapACSE, which is an adaptation of Eq.\ref{eq:arccos}. Instead of fixing margin $m_c$ for all samples in the data, we adapt it to be flexible depending on its negative samples. This new objective improves sentence representation learning by an adaptive margin, $m_c\Delta _{i,j}$, between negative pairs $h_i$ and $h_j$. Thus, if the $\Delta_{i,j}$ is large and the two samples $i,j$ are much different, the margin is also large to push the representation away, and vice versa. 
Our approach is illustrated in the right part of Figure~\ref{fig:kdmcse},
we formalize this objective function as \textbf{AdapACSE}:


\begin{equation}
\begin{split}
& \ell_i^\mathrm{AdapACSE} = \\ -\mathrm{log}  & \frac{ e^{\phi\left ( \theta _{i,i^{*}} \right )/ \tau}}{e^{\phi\left ( \theta _{i,i^{*}} \right )/ \tau} + \sum_{j\neq i}^{n} e^{\phi\left ( \theta _{i,j} - m_c\Delta _{i,j} \right  )/ \tau}}
\end{split}
\end{equation}

\noindent where $\Delta _{i,j} = \left | 1 - \alpha_{i,j} \right |$ is the cosine distance between two samples $i$ and $j$.
Together with the threshold filtering contrastive learning, we rewrite the AdapACSE objective to be:


\begin{equation}
\begin{split}
\label{eq:finaladapcse}
& \ell_i^\mathrm{AdapACSE'} = \ell_i^{m'} = \\  -\mathrm{log}  & \frac{ e^{\phi\left ( \theta _{i,i^{*}} \right )/ \tau}}{e^{\phi\left ( \theta _{i,i^{*}} \right )/ \tau} + \sum_{j\neq i}^{n} \varphi^{t,m}_{i,j} e^{\phi\left ( \theta _{i,j} - m_c\Delta _{i,j} \right  )/ \tau}}
\end{split}
\end{equation}

Finally, we define our KDMCSE objective as follows:

\begin{equation}
    \ell_i^\mathrm{KDMCSE} = \frac{\ell_i^\mathrm{v'} + \ell_i^\mathrm{t'}}{2}
\end{equation}

\noindent where $\ell_i^\mathrm{v'}$ and $\ell_i^\mathrm{v'}$ are AdapACSE Eq.~\ref{eq:finaladapcse} with $t,v$ are text and visual modalities, respectively.


\input{sections/table1}


%% file: sections/table1.tex
\begin{table*}[t]
 \begin{center}
 \scalebox{0.75}{
  \begin{tabular}{clccccccc|c}
    \toprule
    &\textbf{Model} & \textbf{STS12} & \textbf{STS13} & \textbf{STS14} & \textbf{STS15} & \textbf{STS16} & \textbf{STS-B} & \textbf{SICK-R} & \textbf{Avg.$\uparrow$} \\
    \midrule
    \midrule
    \parbox[t]{2mm}{\multirow{2}{*}{\rotatebox[origin=c]{90}{\textit{wiki}}}}
    &SimCSE-BERT$^\diamondsuit$ & 
    67.8$_{\pm 1.6}$ &
    80.0$_{\pm 2.1}$ &
    72.5$_{\pm 1.7}$ &
    80.1$_{\pm 0.8}$ &
    77.6$_{\pm 0.8}$ &
    76.5$_{\pm 0.8}$ &
    70.1$_{\pm 0.9}$ & 
    74.9$_{\pm 1.1}$ \\ 
    &SimCSE-RoBERTa$^\diamondsuit$ &
    68.7$_{\pm1.0}$ &
    82.0$_{\pm0.5}$ &
    74.0$_{\pm1.0}$ &
    82.1$_{\pm0.4}$ &
    81.1$_{\pm0.4}$ &
    80.6$_{\pm0.3}$ &
    69.2$_{\pm0.2}$ &
    76.8$_{\pm0.5}$\\
    \midrule
    \parbox[t]{2mm}{\multirow{6}{*}{\rotatebox[origin=c]{90}{\textit{wiki+flickr}}}}  &SimCSE-BERT$^\diamondsuit$ & 
    69.9$_{\pm 1.7}$ &
    79.8$_{\pm 1.5}$ &
    72.9$_{\pm 0.9}$ &
    81.9$_{\pm 0.8}$ &
    77.8$_{\pm 0.9}$ &
    76.6$_{\pm 1.1}$ &
    68.4$_{\pm 0.8}$ &
    75.3$_{\pm 0.9}$  \\ 
    &\cellcolor{Color}MCSE-BERT$^\diamondsuit$ & 
    \cellcolor{Color}71.4$_{\pm 0.9}$ &
    \cellcolor{Color}81.8$_{\pm 1.3}$ &
    \cellcolor{Color}74.8$_{\pm 0.9}$ &
     \cellcolor{Color}83.6$_{\pm 0.9}$ &
     \cellcolor{Color}77.5$_{\pm 0.8}$ &
     \cellcolor{Color}79.5$_{\pm 0.5}$ &
     \cellcolor{Color}72.6$_{\pm 1.4}$ &
     \cellcolor{Color}77.3$_{\pm 0.5}$  \\ 

    & \cellcolor{Color1}KDMCSE-BERT &
     \cellcolor{Color1}\textbf{74.4}$^*_{\pm 1.4}$ &
     \cellcolor{Color1}\textbf{83.1}$^*_{\pm 0.9}$ &
     \cellcolor{Color1}\textbf{76.3}$^*_{\pm 1.1}$ &
     \cellcolor{Color1}\textbf{83.7}$_{\pm 0.8}$ &
     \cellcolor{Color1}\textbf{78.8}$^*_{\pm 0.9}$ &
     \cellcolor{Color1}\textbf{81.3}$^*_{\pm 0.9}$ &
     \cellcolor{Color1}\textbf{73.0}$^*_{\pm 0.9}$ &
     \cellcolor{Color1}\textbf{78.6}$^*_{\pm 0.8}$ \\ 
     
    \cmidrule{2-10}
    &SimCSE-RoBERTa$^\diamondsuit$  &
    69.5$_{\pm0.9}$&
    81.6$_{\pm0.5}$&
    74.1$_{\pm0.6}$&
    82.4$_{\pm0.3}$&
    80.9$_{\pm0.5}$&
    79.9$_{\pm0.3}$&
    67.3$_{\pm0.5}$&
    76.5$_{\pm0.4}$ \\ 
    &  \cellcolor{Color}MCSE-RoBERTa$^\diamondsuit$  &
     \cellcolor{Color}71.7$_{\pm0.2}$ &
     \cellcolor{Color}82.7$_{\pm0.4}$&
     \cellcolor{Color}75.9$_{\pm0.3}$&
     \cellcolor{Color}\textbf{84.0}$_{\pm0.4}$&
     \cellcolor{Color}81.3$_{\pm0.3}$&
     \cellcolor{Color}\textbf{82.3}$_{\pm0.5}$&
     \cellcolor{Color}70.3$_{\pm1.3}$&
    \cellcolor{Color}78.3$_{\pm0.1}$ \\ 

    &  \cellcolor{Color1}KDMCSE-RoBERTa &
     \cellcolor{Color1}\textbf{73.6}$^*_{\pm 0.7}$&
     \cellcolor{Color1}\textbf{83.8}$^*_{\pm 0.6}$&
     \cellcolor{Color1}\textbf{77.4}$^*_{\pm 0.4}$&
     \cellcolor{Color1}\textbf{84.0}$_{\pm 0.3}$&
     \cellcolor{Color1}\textbf{81.5}$_{\pm 0.7}$&
     \cellcolor{Color1}\textbf{82.3}$_{\pm 0.6}$&
     \cellcolor{Color1}\textbf{71.2}$^*_{\pm 0.4}$&
     \cellcolor{Color1}\textbf{79.1}$^*_{\pm 0.3}$ \\
    
    \midrule
    \parbox[t]{2mm}{\multirow{6}{*}{\rotatebox[origin=c]{90}{\textit{wiki+coco}}}}&SimCSE-BERT$^\diamondsuit$  &  
    69.1$_{\pm 1.0}$ &
    80.4$_{\pm 0.9}$ &
    72.7$_{\pm 0.7}$ &
    81.1$_{\pm 0.3}$ &
    78.2$_{\pm 0.9}$ &
    73.9$_{\pm 0.6}$ &
    66.6$_{\pm 1.2}$ &  
    74.6$_{\pm 0.2}$ \\ 
    & \cellcolor{Color}MCSE-BERT$^\diamondsuit$ &
     \cellcolor{Color}71.2$_{\pm 1.3}$ &
     \cellcolor{Color}79.7$_{\pm 0.9}$ &
     \cellcolor{Color}73.8$_{\pm 0.9}$ &
     \cellcolor{Color}83.0$_{\pm 0.4}$ &
     \cellcolor{Color}77.8$_{\pm 0.9}$ &
     \cellcolor{Color}78.5$_{\pm 0.4}$ &
     \cellcolor{Color}72.1$_{\pm 1.4}$ &
     \cellcolor{Color}76.6$_{\pm 0.5}$ \\ 

    & \cellcolor{Color1}KDMCSE-BERT &
     \cellcolor{Color1}\textbf{73.2}$^*_{\pm 1.2}$ &
     \cellcolor{Color1}\textbf{80.5}$^*_{\pm 1.0}$ &
     \cellcolor{Color1}\textbf{75.4}$^*_{\pm 0.9}$ &
     \cellcolor{Color1}\textbf{83.2}$_{\pm 0.3}$ &
     \cellcolor{Color1}\textbf{79.7}$^*_{\pm 0.8}$ &
     \cellcolor{Color1}\textbf{79.7}$^*_{\pm 0.7}$ &
     \cellcolor{Color1}\textbf{73.7}$^*_{\pm 1.4}$ &
     \cellcolor{Color1}\textbf{77.9}$^*_{\pm 1.2}$ \\ 
    \cmidrule{2-10}
    & SimCSE-RoBERTa$^\diamondsuit$ &
    66.4$_{\pm 0.9}$&
    80.7$_{\pm 0.7}$&
    72.7$_{\pm 1.1}$&
    81.3$_{\pm 0.9}$&
    80.2$_{\pm 0.8}$&
    76.8$_{\pm 0.6}$&
    65.7$_{\pm 0.7}$&
    74.8$_{\pm 0.5}$ \\ 
    &  \cellcolor{Color}MCSE-RoBERTa$^\diamondsuit$ &
     \cellcolor{Color}70.2$_{\pm 1.7}$&
     \cellcolor{Color}\textbf{82.0}$_{\pm 0.7}$&
     \cellcolor{Color}75.5$_{\pm 1.2}$&
     \cellcolor{Color}83.0$_{\pm 0.6}$&
     \cellcolor{Color}\textbf{81.5}$_{\pm 0.7}$&
     \cellcolor{Color}\textbf{80.8}$_{\pm 1.0}$&
     \cellcolor{Color}\textbf{69.9}$_{\pm 0.6}$&
     \cellcolor{Color}77.6$_{\pm 0.8}$ \\

    &  \cellcolor{Color1}KDMCSE-RoBERTa &
     \cellcolor{Color1}\textbf{72.8}$^*_{\pm 1.5}$&
     \cellcolor{Color1}81.7$_{\pm 0.9}$&
     \cellcolor{Color1}\textbf{76.1}$^*_{\pm 1.1}$&
     \cellcolor{Color1}\textbf{83.4}$^*_{\pm 1.0}$&
     \cellcolor{Color1}\textbf{81.5}$_{\pm 0.6}$&
     \cellcolor{Color1}80.7$_{\pm 0.8}$&
     \cellcolor{Color1}\textbf{69.9}$_{\pm 0.6}$&
     \cellcolor{Color1}\textbf{78.0}$^*_{\pm 0.7}$ \\
  \bottomrule
  \end{tabular}}\\
  \caption{Performance comparison on STS tasks. STS-B: STS Benchmark, SICK-R: SICK-Relatedness, Avg.: average across $7$ tasks. $^\diamondsuit$ : results from \cite{zhang2022mcse}. We train the models using a random seed of $5$, presenting the average and standard deviations of our findings. $\ast$: difference between MCSE and KDMCSE is significant at $\alpha=0.05$ according to an independent t-test. }
  \vspace{-5pt}
  \label{tab:wiki}
  \end{center}
\end{table*}

%% file: sections/experiments.tex
\section{Experiments Setup}

\subsection{Dataset}

We employ Flickr30k~\citep{young2014flickr} and MS-COCO~\citep{lin2014coco} as our multimodal datasets. Flickr30k contains $29,783$ training images and MS-COCO has $82,783$ training images. Every image in these collections comes with multiple captions, usually five captions. As in \citet{gao2021simcse}, we utilize Wiki1M for our text-based dataset, consisting of $10^6$ sentences extracted randomly from English Wikipedia.

\subsection{Implementation}
\paragraph{Language Encoder - Student model} 

We have implemented our work in the Hugging Face Transformers library\footnote{\href{https://github.com/huggingface/transformers}{https://github.com/huggingface/transformers}} as described by \citep{wolf-etal-2020-transformers}. We load our language encoder from the checkpoints of \texttt{bert-base-uncased} and \texttt{roberta-base}. Then, we fine-tune these foundation models with our introduced contrastive objective. To evaluate sentence embeddings, we use the 768-dimensional outputs of the \texttt{[CLS]} token preceding the MLP pooler layer from the transformer-based models.

\paragraph{Multimodal encoder - Teacher model}

Our teacher model is CLIP~\cite{radford2021clip}, the model's weights are initialized with pre-trained \texttt{clip-vit-base-patch32}, where \texttt{ViT-B/32 Transformer} architecture is built as an image encoder and the patch size $P$ is 32. Because during training, the teacher model will not be fine-tuned, we will pre-extract image and text features $\hat{\vv},\hat{\vt}$ to reduce the computation cost during the training phase.

\paragraph{MLP Projection Heads} 

We use 4 different MLP modules $g_{\{\theta_1,\theta_2,\theta_t,\theta_v\}}$ to project different modalities and objectives. For the pure textual objective using the Wiki1M dataset, sentence embeddings are projected into a 768-dimensional space. For CLIP projection, both the sentence embeddings (student and teacher models) and the image feature vectors are projected into a shared 256-dimensional space.

\paragraph{Parameter Settings}


In line with the experimental setup described by \citet{zhang2022mcse}, we explored two primary training scenarios: \textit{wiki+flickr} and \textit{wiki+coco}. Mini-batches were drawn either from the Wiki1M corpus or from the respective caption datasets, with proportions reflecting their relative sizes. Our parameter choices were largely inspired by the guidelines provided by \citet{gao2021simcse}. Specifically, we set the temperature parameters $\tau$ and $\tau'$, to $0.05$. For model assessment, we performed evaluations on the STS-B dev set after every $125$ training iteration, preserving the best performing checkpoint for our final evaluation. Regarding the BERT encoder, we adopted a learning rate of $3e-5$ and a batch size of $64$. For RoBERTa, these parameters were adjusted to $1e-5$ and $128$, respectively. Our training was executed on an A6000 GPU, with each experimental run spanning roughly 5-6 hours and equivalent to MCSE training time.

 \begin{table}[t]
 \begin{center}
 \scalebox{0.75}{
  \begin{tabular}{l|cc|cc}
    \toprule
    \multirow{2.5}{*}{\textbf{Model}} & \multicolumn{2}{c}{\textit{alignment}$\downarrow$} & \multicolumn{2}{c}{\textit{uniformity}$\downarrow$}\\
    \cmidrule{2-5}

     &  \textit{flickr} & \textit{coco} & \textit{flickr} & \textit{coco} \\
    \midrule
    MCSE-BERT & 0.293 & 0.267 & \textbf{-2.491} & -2.350 \\
    KDMCSE-BERT & \textbf{0.245} & \textbf{0.261} & -2.387 & \textbf{-2.383} \\
    \midrule
    MCSE-RoBERTa & 0.209 & 0.195 & -1.721 & -1.418 \\
    KDMCSE-RoBERTa & \textbf{0.174} & \textbf{0.149} & \textbf{-1.952} & \textbf{-1.748} \\
    \bottomrule
  \end{tabular}}
  \vspace{-5pt}

  \caption{ The alignment uniformity results of the models when using the BERT and RoBERTa encoder. All models are trained in the \textit{wiki-flickr} setting.}
  \label{tab:ali_uni}
  \end{center}
\end{table}


\subsection{Evaluation}

We evaluate our trained models on seven STS (Semantic Textual Similarity) tasks, including STS 2012-2016~\citep{agirre2012semeval, agirre2013sem, agirre2014semeval, agirre2015semeval, agirre2016semeval}, the STS Benchmark~\citep{cer2017semeval}, and SICK-Relatedness~\citep{marelli2014sick}. Each dataset is made up of pairs of sentences with the aim of assigning a similarity score to each pair. In alignment with \citet{gao2021simcse}, we present the Spearman correlation (multiplied by 100) between the official annotations and our predicted scores in an "all" context. This means that we merge all subsets for each task and then provide a comprehensive Spearman correlation.

\section{Experiments Results}

\subsection{Main Results}


Following the methodology of previous studies, we conducted our model five times using the same hyperparameter settings. Averaged results are delineated in Table~\ref{tab:wiki}. A collective overview of our results reveals that our model consistently matches or exceeds the benchmarks set by baseline approaches. MCSE advances beyond SimCSE in the majority of STS evaluations. By integrating auxiliary visual and textual insights derived from the CLIP teacher model, our KDMCSE model showcases pronounced improvements in various downstream tasks. Specifically, when trained on Wiki1M and Flickr30k datasets, KDMCSE increases the performance metrics for BERT (from $77.3$ to $78.6$) and RoBERTa (from $78.3$ to $79.1$). Compared to MCSE-RoBERTa, our model exhibits comparable metrics on STS15 and STS-B, with a slight advancement on STS16. In the context of the \textit{wiki+coco} dataset, KDMCSE-BERT notably surpasses other methods across most tasks, with scores rising from $76.6$ to $77.9$ (STS15 being the exception with a modest increase). Our RoBERTa encoder, when trained with the KDMCSE framework, showed higher average results (an increase from $77.6$ to $78.0$), where tasks such as STS12, STS14, and STS15 registered significant improvements, while others retained or slightly underperformed in comparison.

\subsection{Alignment and Uniformity}

The concepts of alignment and uniformity are closely linked to contrastive learning, serving as potential metrics to evaluate the quality of the representations derived~\cite{wang2020understanding}. Alignment is a property that encourages encoders to produce analogous representations for instances that are alike. This can be quantified using the expected distance between the embeddings of positively paired instances.

\vspace{-4pt}
\begin{equation}
    \mathcal{L}_{align}\triangleq \mathop{\mathbb{E}}_{(x, x^+)~\sim~ p_{\text{pos}}}\left\| f(x)-f(x^+) \right\|^2_2 \; .
\end{equation}
\vspace{-4pt}

And the \textit{uniformity loss} prefers a uniform distribution in the hypersphere:

\vspace{-4pt}
\begin{equation}
    \mathcal{L}_{uniform}\triangleq \log \mathop{\mathbb{E}}_{x,y \stackrel{i.i.d.}{\sim} p_{\text{data}}} e^{-2\left\| f(x)-f(y) \right\|^2_2} \;.
\end{equation}
\vspace{-4pt}

To delve deeper into the mechanics of our methodology, we assess alignment and uniformity metrics using the STS-B development set. In our comparison with MCSE, the results are presented in Table~\ref{tab:ali_uni}. Evidently, when compared to MCSE for both BERT and RoBERTa-based encoders, KDMCSE shows superior performance in both alignment and uniformity metrics. This not only reaffirms the foundational rationale of our method, but also suggests that AdapACSE can effectively enhance the caliber of sentence embeddings.

%% file: sections/analysis.tex
\section{Analysis}



\subsection{Exploring the Impact of Angular Margin}



\begin{figure}[t]
\centering
\includegraphics[width=0.5\textwidth] {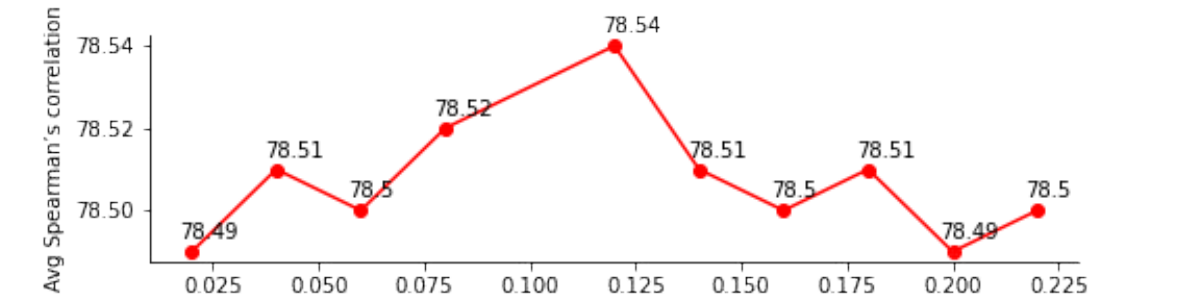}
\caption{Impact of angular margin $m$ in AdapACSE. The results are reported on average across 7 tasks, and our experiments are conducted on BERT encoder and in \textit{wiki-flickr} setting.  }
\vspace{-5pt}
\label{fig:vismargin}
\end{figure}

In our AdapACSE loss function, the angular margin, represented as $m$, is critical to determining the discriminative capability. We conducted an experiment to better understand the impact of $m$ by varying it in steps of 0.025 radians from 0.025 radians to 0.225 radians. The performance metrics averaged on all tasks are depicted in Figure~\ref{fig:vismargin}. From the data, we discern an optimal performance when $m$ is set to 0.125. Any deviation from this optimal point, be it an increase or decrease, results in a performance drop. This aligns with our preliminary hypotheses: A smaller value of $m$ could have a negligible effect, whereas an overly large $m$ could negatively skew the representations within the fusion space.

\begin{figure}[t]
\centering
\includegraphics[width=0.5\textwidth] {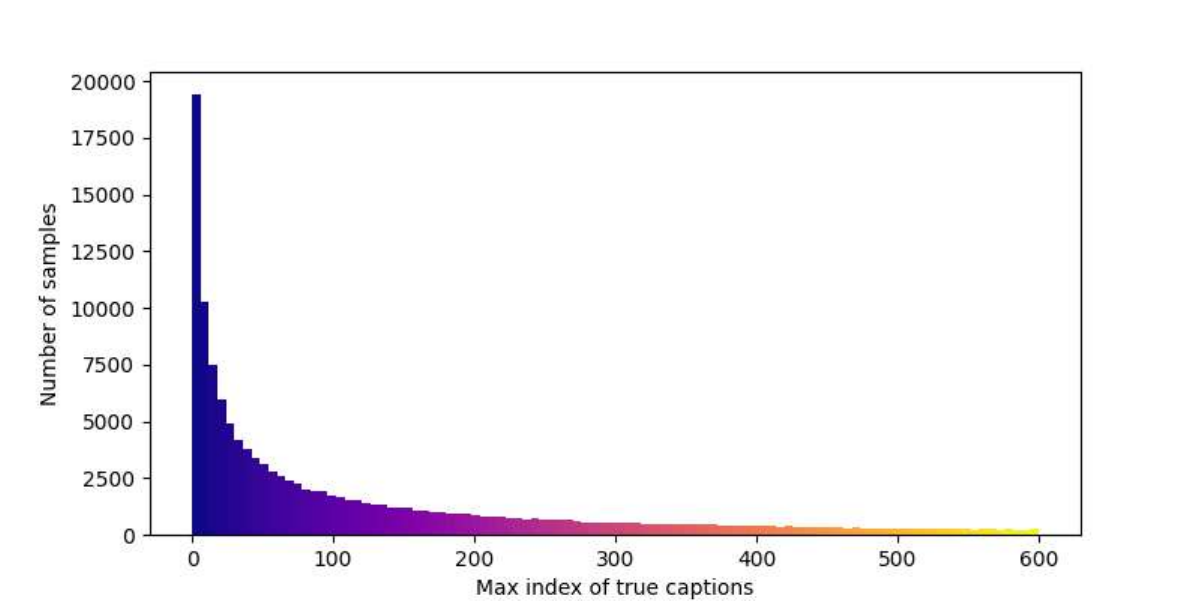}
\caption{Statistic of the maximum index of true captions when sorting the similarity score between image and text of Flickr dataset.}
\vspace{-5pt}
\label{fig:stat}
\end{figure}

\begin{figure}[t]
\centering
\includegraphics[width=0.5\textwidth] {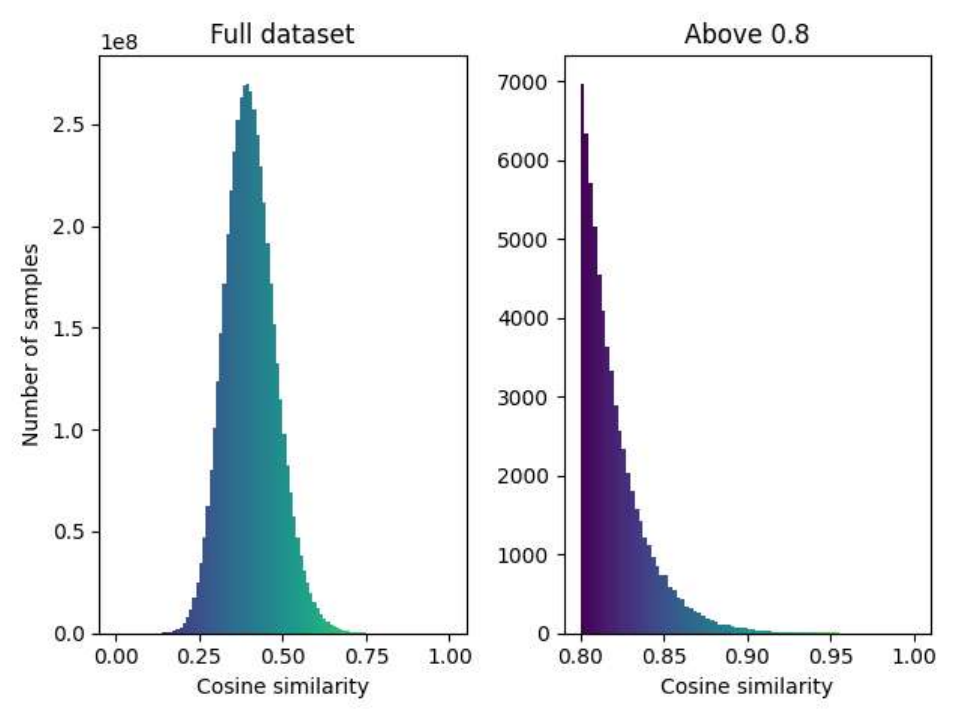}
\caption{Statistic of the similarity score between image and text of Flickr dataset.}
\vspace{-5pt}
\label{fig:threshold}
\end{figure}

\subsection{Threshold Filtering selection}

In this section, we will show how to define the threshold hyperparameter during training. We make statistics based on text-text and text-visual similarity scores in the Flickr set, and select based on that statistic. We perform two calculations: (1) the histogram for the similarity threshold and (2) the position of the largest true captions (the annotation captions of that image) that align with the image when sorted with all captions in the dataset with that image. Looking at Figure~\ref{fig:stat}, we see that most of the true captions of an image are in the top 100 (out of about 150,000 captions). However, there are still quite a few captions that, even though they accurately describe that image when sorted, can still be pushed to very far indexes (>200). This may be because there are still many similar images in the dataset, leading to captions that describe another image, but are still correct for this image. The image~\ref{fig:threshold} shows that the scores are in a normal distribution. We focus more clearly on samples with high similarity and see that there are only a few pairs with similarity higher than 0.8. and decreases rapidly as it approaches 1.0. Based on the above statistics, we have chosen a threshold of about 0.85-0.9 to eliminate pairs in the contrastive learning objective.

%% file: sections/conclusion.tex
\section{Conclusion}
In this paper, we introduce KDMCSE, a novel approach to sentence embedding learning that transfers knowledge from a vision and language model with a multimodal contrastive objective to not only align sentences and the corresponding image and text representations of the teacher model but also avoid noisy negative samples. Additionally, our innovative contrastive objective, AdapACSE, addresses the challenges associated with capturing nuanced differences within negative pairs, strengthening discriminative representation. Our framework lays the groundwork for continued exploration in this domain, fostering advancements in vision and language applications.


%% file: sections/limitations.tex
\section{Limitations}

In this study, we shed light on the promising enhancements of KDMCSE in STS benchmarks. However, it is imperative to acknowledge its limitations. In particular, there is a pronounced disparity in both the distribution and volume of word tokens when visual datasets are compared to traditional language corpora. Taking the Book Corpus and Wikipedia as instances, these standard language repositories house billions of words, spanning millions of unique tokens. In contrast, MS COCO, a frequently cited visual-grounded dataset, encompasses merely a million words and about a thousand unique tokens. Another challenge encountered in our methodology is the intricate correlation of the hyperparameter in our contrastive objective, which remains elusive. Its optimal configuration was determined not through systematic understanding, but rather through exhaustive fine-tuning across many experiments. Furthermore, the characterization of "semantic similarity" is highly dependent on the specific task at hand. Although we have seen significant advancements in STS benchmarks, there remains a need to evaluate the disparity in performance between text-only models and their multimodal counterparts across diverse benchmarks. This would provide a more holistic understanding of the efficacy of sentence representations in varied contexts.

%% file: sections/appendix.tex
\onecolumn

\input{sections/table2}

\begin{table*}[ht]
 \begin{center}
 \scalebox{0.75}{
  \begin{tabular}{lccccccc|c}
    \toprule
    \textbf{Model} & \textbf{STS12} & \textbf{STS13} & \textbf{STS14} & \textbf{STS15} & \textbf{STS16} & \textbf{STS-B} & \textbf{SICK-R} & \textbf{Avg.$\uparrow$} \\
    \midrule
    \midrule

    \multicolumn{9}{c}{Subset 1}\\ 
    \midrule
    MCSE&
     71.54&
     77.9&
     73.6&
     80.83&
     78.38&
     79.73&
     74.66&
     76.66
\\
    KDMCSE &
     73.00&
     78.53&
     74.93&
     82.14&
     79.88&
     81.37&
     74.08&
     77.7
\\
    \bottomrule
 \multicolumn{9}{c}{Subset 2}\\
 \midrule
 MCSE
& 71.67& 78.56& 74.02& 81.19& 78.54& 80.29& 73.87&76.88
\\
 KDMCSE & 73.35& 79.51& 75.58& 83.43& 80.07& 81.89& 73.78&78.23
\\
 \end{tabular}}\\
   \vspace{1mm}

  \caption{Ablation results on STS tasks. STS-B: STS Benchmark, SICK-R: SICK-Relatedness, Avg.: average across $7$ tasks. The models are trained with the BERT encoder and in 1M image-text pairs subsets of the CC12 dataset.}
  \label{tab:re_cc12}
  \end{center}
\end{table*}

\begin{table*}[ht]
 \begin{center}
 \scalebox{0.75}{
  \begin{tabular}{lccccccc|c}
    \toprule
    \textbf{Model} & \textbf{STS12} & \textbf{STS13} & \textbf{STS14} & \textbf{STS15} & \textbf{STS16} & \textbf{STS-B} & \textbf{SICK-R} & \textbf{Avg.$\uparrow$} \\
    \midrule
    MCSE-BERT$_{CLIP Text}$& 
    72.60&
    81.86&
    75.28&
    83.28&
    78.74&
    80.77&
    72.78&
    77.90\\ 
    KDMCSE-BERT& 
    74.40&
    83.10&
    76.26&
     83.68&
     78.79&
     81.25&
     73.03&
     78.64
\\
    
    \midrule
    MCSE-BERT$_{CLIP Text}$& 
    71.39&
    79.25&
    73.84&
    82.65&
    77.96&
    78.85&
    71.74&  
    76.53
\\ 
    KDMCSE-BERT&
     73.38&
     80.71&
     75.55&
     83.36&
     79.91&
     79.85&
     73.92&
     78.10\\ 
  \end{tabular}}\\
  \caption{Ablation study on STS tasks. STS-B: STS Benchmark, SICK-R: SICK-Relatedness, Avg.: average across $7$ tasks.. We train the MCSE models using additional CLIP text embeddings.  }
  
  \label{tab:mcse_with_clip_text}
  \end{center}
\end{table*}

\section{Regularization}

To validate the significance of our objectives, we select the best-performing models for each dataset and gradually run with three settings: our KDMCSE, KDMCSE without AdapACSE, and KDMCSE without threshold filtering. The results are presented in Table \ref{tab:re2}, indicating that the incorporation of all the proposed approaches leads to the best performance in all tasks. The second line shows that not applying threshold filtering can marginally harm performance. This result can be attributed to the fact that, during training, noisy negative samples do not often appear in the batch. In the last row of the table, we show that the removal of AdapACSE significantly affects the performance of both BERT and RoBERTa.

\section{Larger experiments on image-text paired data}

To mitigate scalability and generalization concerns, we conducted extensive experiments using a significantly larger dataset containing one million image-text pairs sourced from the CC12 dataset, equivalent in sample size to Wiki1M. In Table~\ref{tab:re_cc12}, we present the results of these experiments. Our findings consistently demonstrate that our KDMCSE model surpasses MCSE in terms of average performance (Avg) across STS, across various subtasks within CC12M subsets.

\section{Impact of Adaptive Angular Margin Contrastive objective}

To substantiate the significance of our objectives and ensure fairness in our comparisons, we carried out experiments involving MCSE with the inclusion of additional text embeddings from the CLIP text encoder. The results of these experiments are presented in Table~\ref{tab:mcse_with_clip_text}. Our findings unequivocally indicate that our proposed contrastive objectives have a positive impact on the model's performance, thereby strengthening our core objectives.

\section{KDMCSE algorithm}

To enhance the presentation of our method, we provide a detailed description of our implementation, adhering to the pseudo code outlined in Algorithm~\ref{algo:kdmcse}.

\begin{algorithm}

\caption{Our KDMCSE algorithm}\label{algo:kdmcse}
\begin{algorithmic}

\STATE {\bfseries Input:} 
\STATE Collection of sentences (text-only dataset) $D=\{x_i\}_{i=1}^n$
\STATE Collection of image-text pairs (multimodal dataset) $D^M=\{ x_i, y_i  \}_{i=1}^{n_m}$ 
\STATE Paired sample step $p = \left \lceil \frac{\left| D\right|}{\left| D^M\right|}\right \rceil$
\FOR{$t=1, 2, 3, \ldots, iter$}
    \IF{not $t | p$}
        \STATE \footnotesize{Sample batch} $\{x_i\}_{bs}$ \footnotesize{from} $D$
        \STATE \footnotesize{// Forward through language model and projection}
        \STATE $\vh_i^{z},\vh_i^{z'}=g_{\phi}(f_{\theta}(x_i))$ 
        \STATE  \footnotesize{// Calculate loss} 
        \STATE $\frac{1}{bs} \sum_{i}^{bs} \ell_i^\mathrm{S}$
    \ELSE
        \STATE \footnotesize{Sample batch} $\{x_i, y_i\}_{bs}$ \footnotesize{from} $D^M$
        \STATE $\vs_i^z = g_{\phi_g}(f_{\theta}(x_i, z))$
        \STATE $\vt_i = g_{\phi_t}(\hat{\vt_i}),\hat{\vt_i} = CLIP_{text}(x_i)$
        \STATE $\vv_i = g_{\phi_v} (\hat{\vv_i}),\hat{\vv_i} = CLIP_{visual}(y_i)$
        \STATE \footnotesize{// Calculate cosine similarity}
        \STATE $\alpha^{t,t}_{i,j} = \dfrac {\vs_i \cdot \vt_j} {\left\| \vs_i\right\| _{2}\left\| \vt_j\right\| _{2}}$
        \STATE $\alpha^{t,v}_{i,j} = \dfrac {\vs_i \cdot \vv_j} {\left\| \vs_i\right\| _{2}\left\| \vv_j\right\| _{2}}$
        \STATE \footnotesize{// Calculate soft-label (text-text, text-visual)}
        \STATE $tt_{score} = \dfrac {\hat{\vt_i} \cdot \hat{\vt_j}} {\left\| \hat{\vt_i}\right\| _{2}\left\| \hat{\vt_j}\right\| _{2}}$
        \STATE $tv_{score} = \dfrac {\hat{\vt_i} \cdot \hat{\vv_j}} {\left\| \hat{\vt_i}\right\| _{2}\left\| \hat{\vv_j}\right\| _{2}} $
        \STATE \footnotesize{// Apply threshold filtering TF based on soft-label}
        \STATE $TF(\alpha^{t,t}_{i,j})$
        \STATE $TF(\alpha^{t,v}_{i,j})$
        \STATE \footnotesize{// Calculate AdapACSE loss} 
        \STATE $\frac{1}{bs} \sum_{i}^{bs} \ell_i^\mathrm{KDMCSE}$
    \ENDIF
\ENDFOR
\end{algorithmic}
\end{algorithm}

\section{Sentence embedding visualization}

\begin{figure}[h!]
\centering
\begin{subfigure}[t]{0.45\linewidth} 
    \centering
  \includegraphics[width=\linewidth]{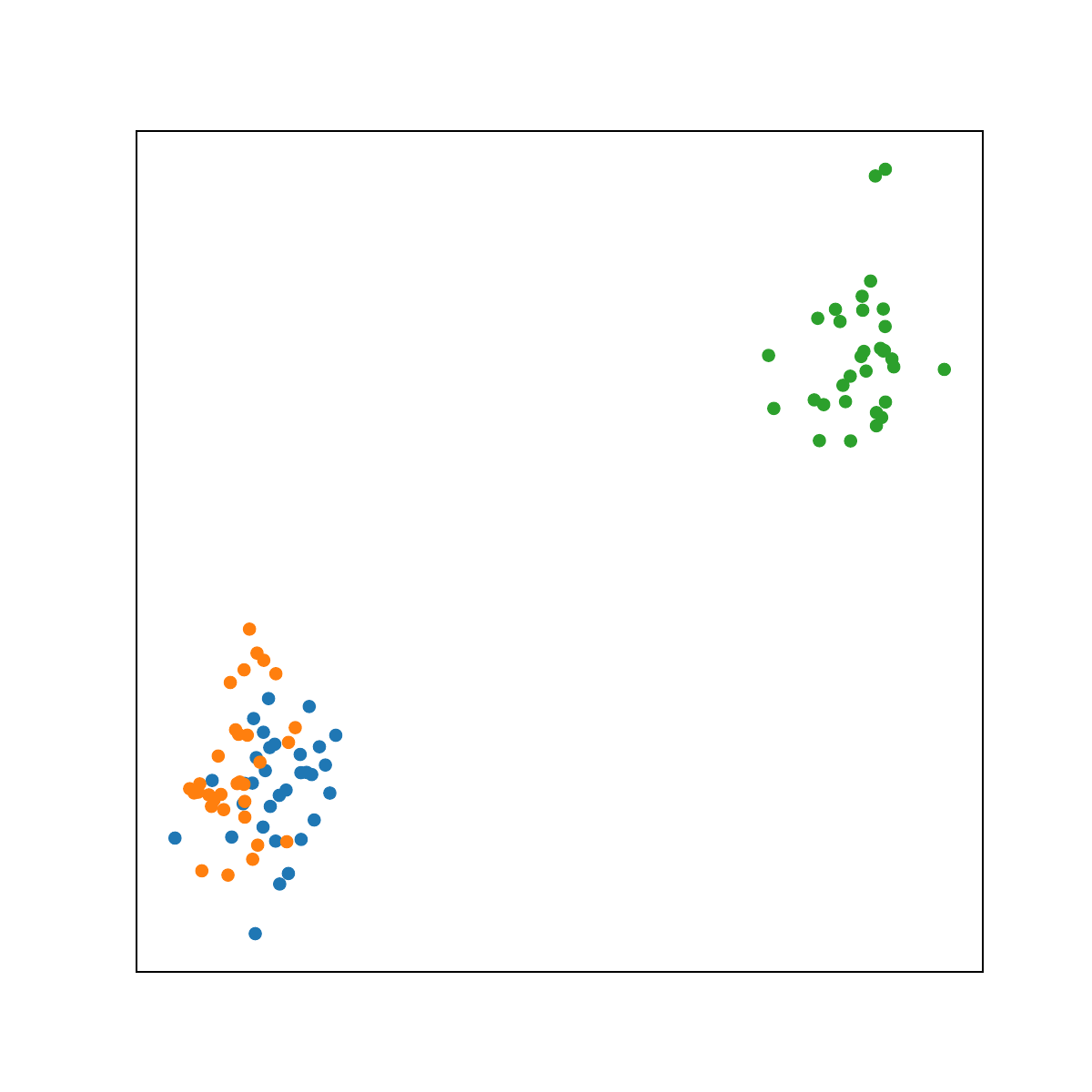}
  \caption{KDMCSE} 
\end{subfigure}
\begin{subfigure}[t]{0.45\linewidth} 
    \centering
  \includegraphics[width=\linewidth]{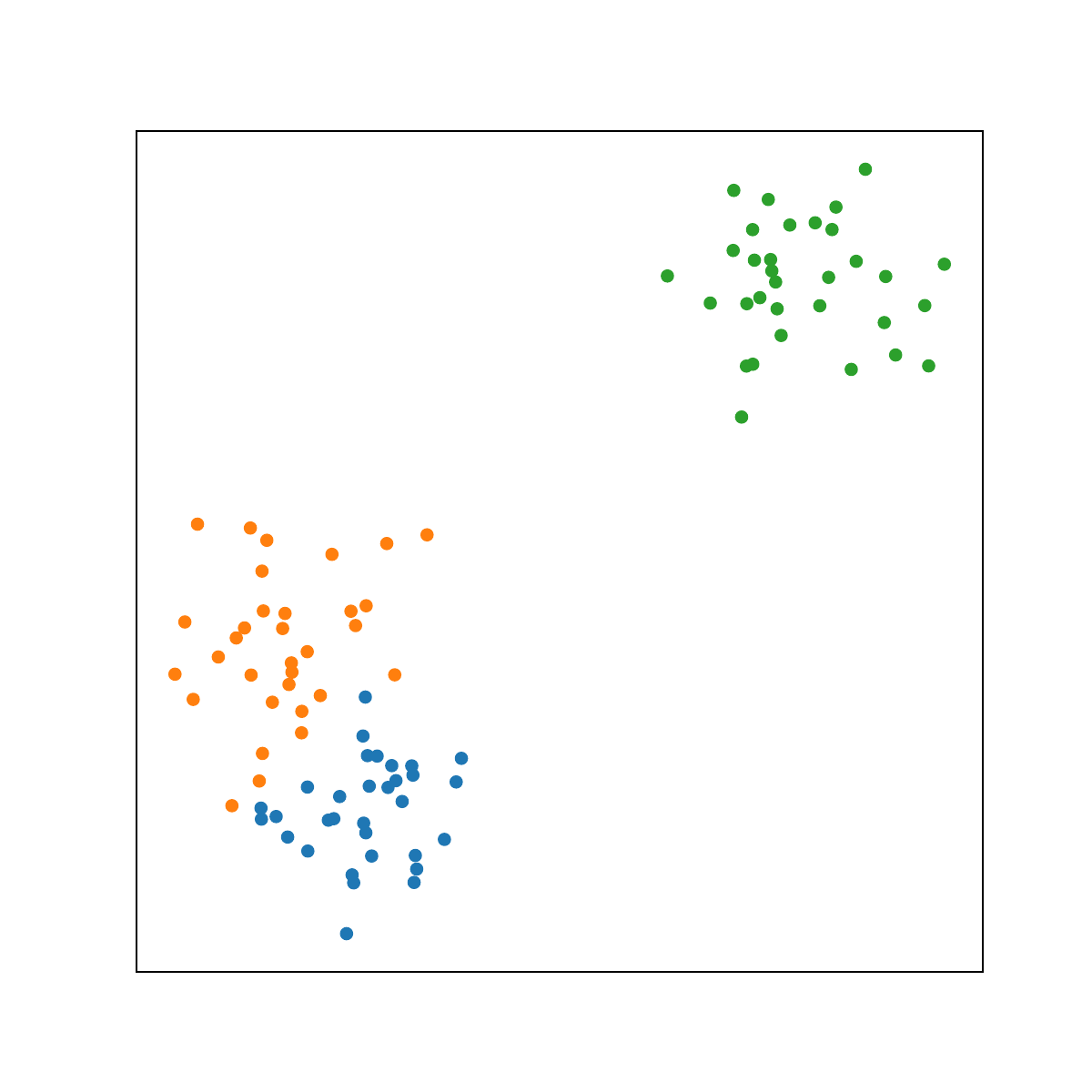}
  \caption{MCSE}
\end{subfigure}
\caption{\textcolor{orange}{Sentence 1}: A white and black dog and a brown dog in sandy terrain. \textcolor{blue}{Sentence 2}: Brown , black and white dog standing on a sandy slope. \textcolor{green}{Sentence 3}: Two girls stand up against a red wall.}
\label{fig:example1}
\end{figure}

\begin{figure}[h!]
\centering
\begin{subfigure}[t]{0.45\linewidth} 
    \centering
  \includegraphics[width=\linewidth]{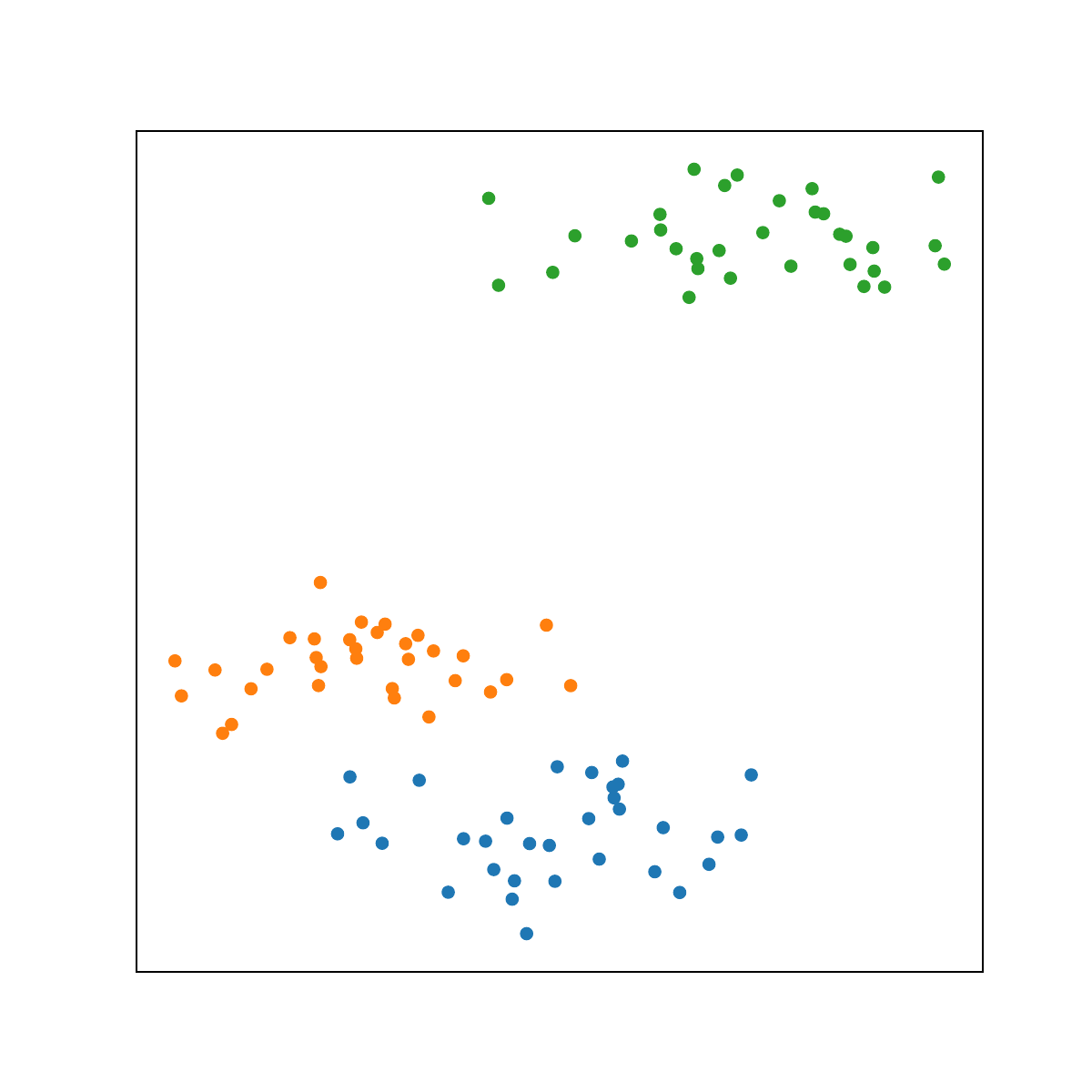}
  \caption{KDMCSE} 
\end{subfigure}
\begin{subfigure}[t]{0.45\linewidth} 
    \centering
  \includegraphics[width=\linewidth]{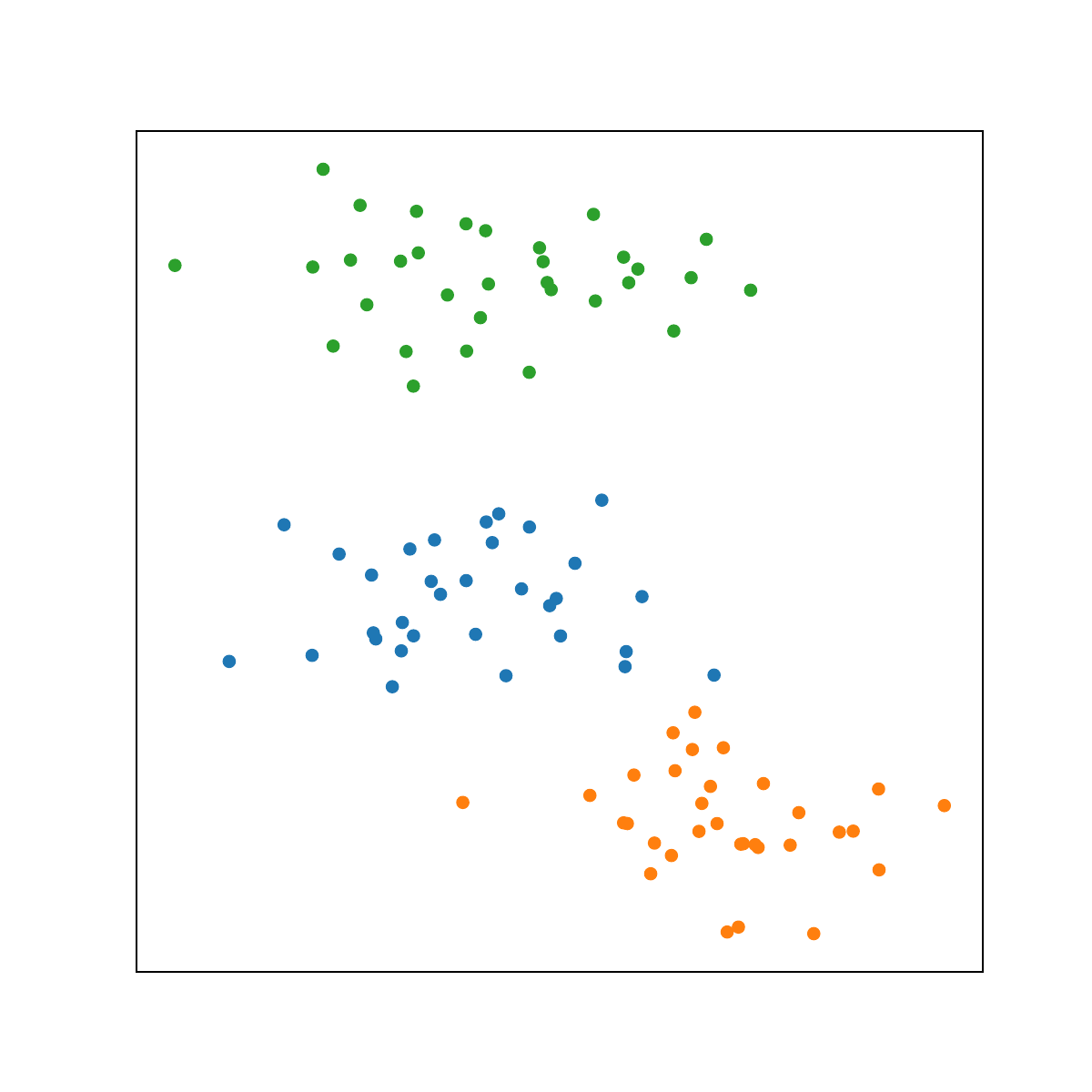}
  \caption{MCSE}
\end{subfigure}
\caption{\textcolor{orange}{Sentence 1}: An elderly man holding a pitchfork and doing some yard work. \textcolor{blue}{Sentence 2}: An old bearded man is tilling the soil with a simple wooden plow , with a wooden fence and an old barn in the background. \textcolor{green}{Sentence 3}: A man in a red shirt is performing an aerial trick with a skateboard on a sidewalk.}
\label{fig:example2}
\end{figure}

\begin{figure}[h!]
\centering
\begin{subfigure}[t]{0.45\linewidth} 
    \centering
  \includegraphics[width=\linewidth]{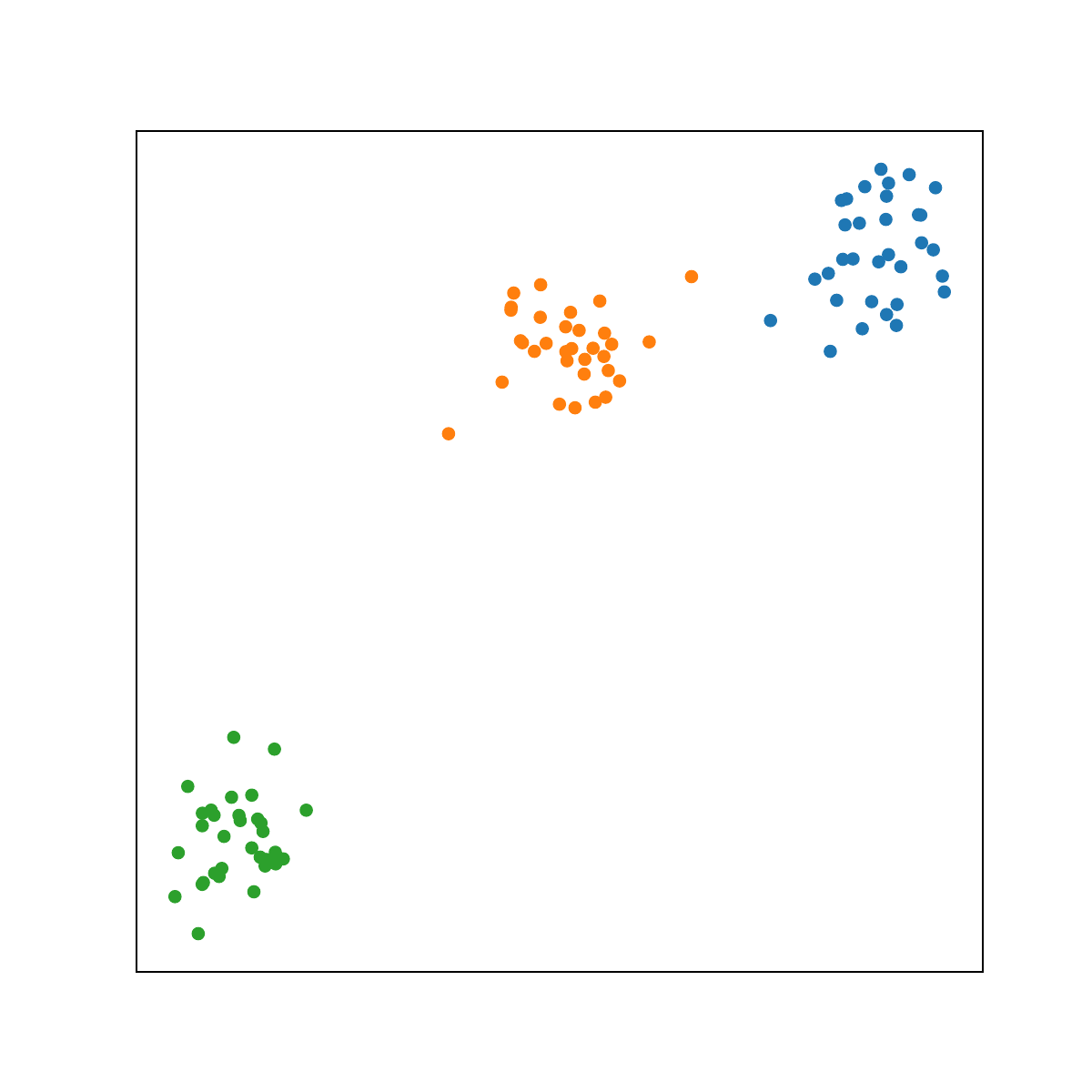}
  \caption{KDMCSE} 
\end{subfigure}
\begin{subfigure}[t]{0.45\linewidth} 
    \centering
  \includegraphics[width=\linewidth]{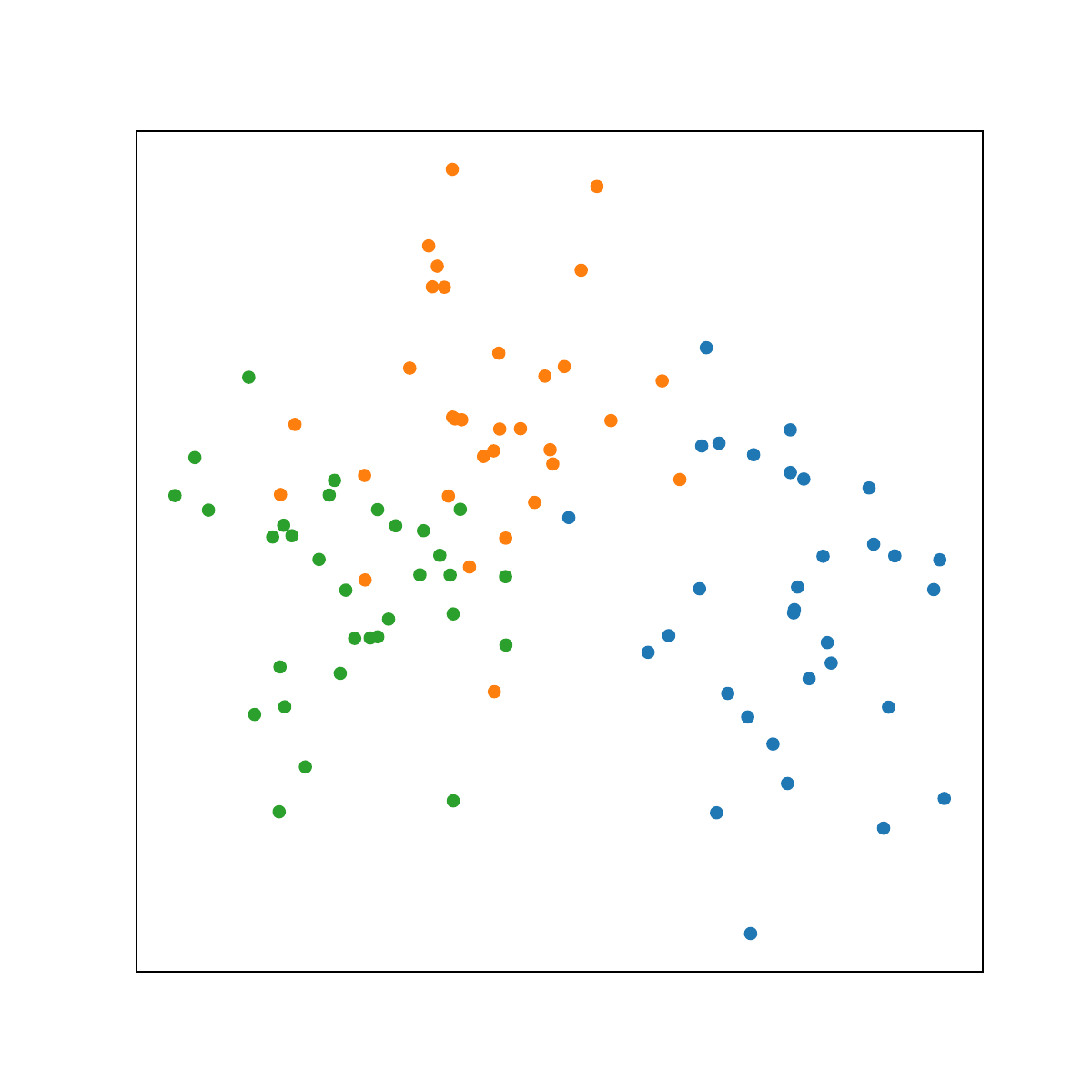}
  \caption{MCSE}
\end{subfigure}
\caption{\textcolor{orange}{Sentence 1}: A person wearing a blue shirt , rides a white horse along a dusty country road. \textcolor{blue}{Sentence 2}: A man in a blue tee-shirt , wearing a cap , is riding a white horse down a dirt road , in a rural setting of grass and hills. \textcolor{green}{Sentence 3}: A man climbing a mountain.}
\label{fig:example3}
\end{figure}

\begin{figure}[h!]
\centering
\begin{subfigure}[t]{0.45\linewidth} 
    \centering
  \includegraphics[width=\linewidth]{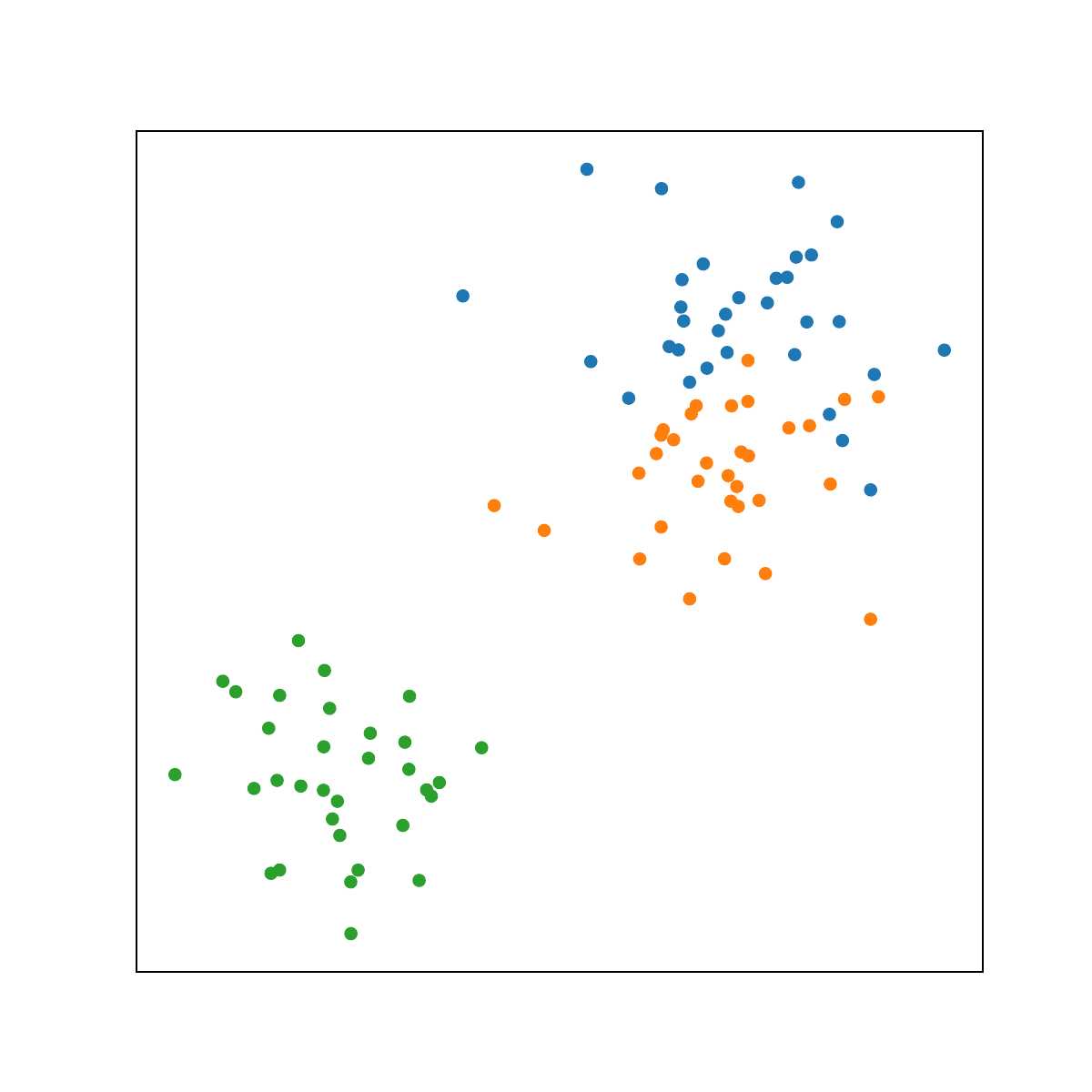}
  \caption{KDMCSE} 
\end{subfigure}
\begin{subfigure}[t]{0.45\linewidth} 
    \centering
  \includegraphics[width=\linewidth]{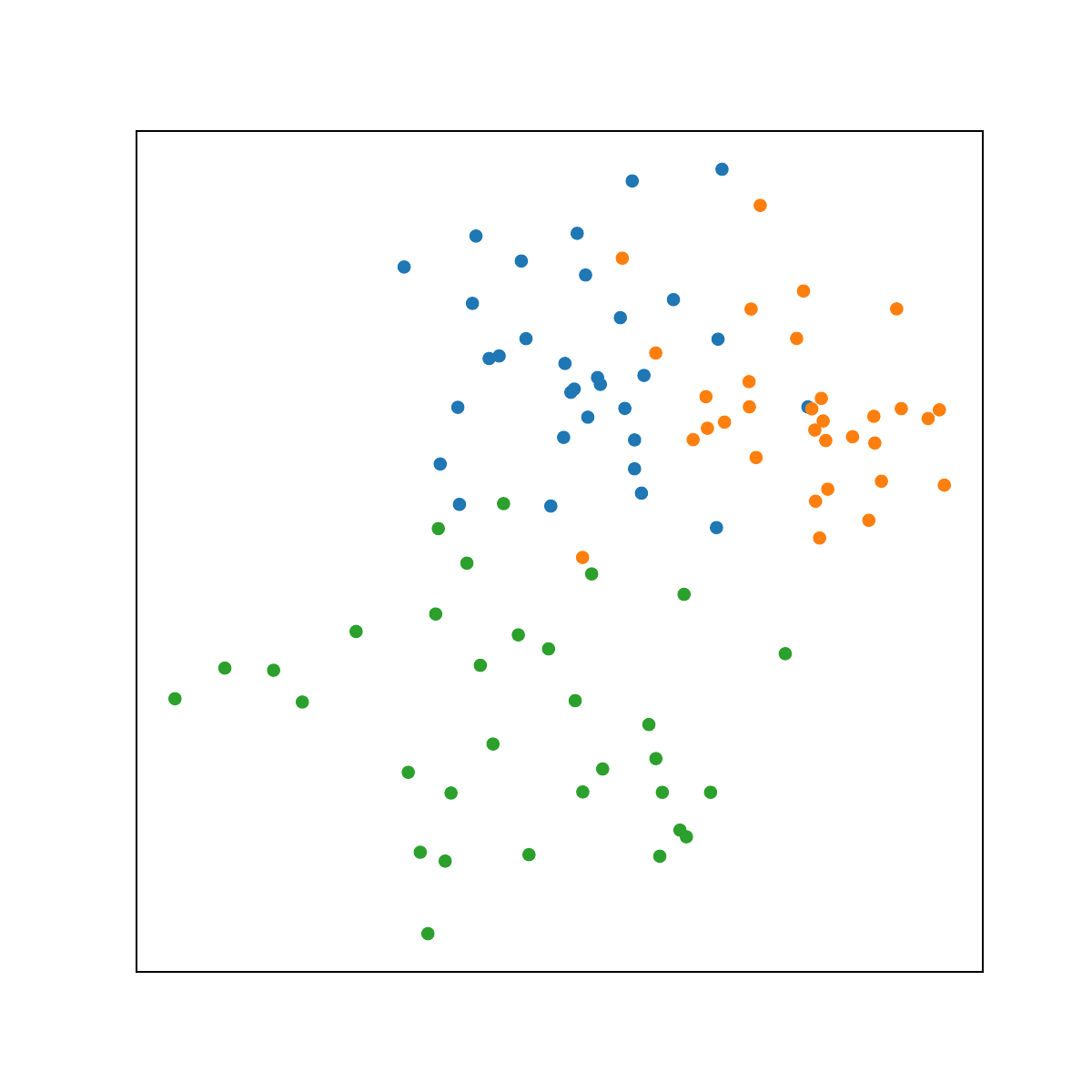}
  \caption{MCSE}
\end{subfigure}
\caption{\textcolor{orange}{Sentence 1}: A dog is jumping into a swimming pool after a duck. \textcolor{blue}{Sentence 2}: A dog is jumping into a pool to get a duck floating there. \textcolor{green}{Sentence 3}: A white curly-haired dog runs with a stick in its mouth.}
\label{fig:example4}
\end{figure}

\begin{figure}[h!]
\centering
\begin{subfigure}[t]{0.45\linewidth} 
    \centering
  \includegraphics[width=\linewidth]{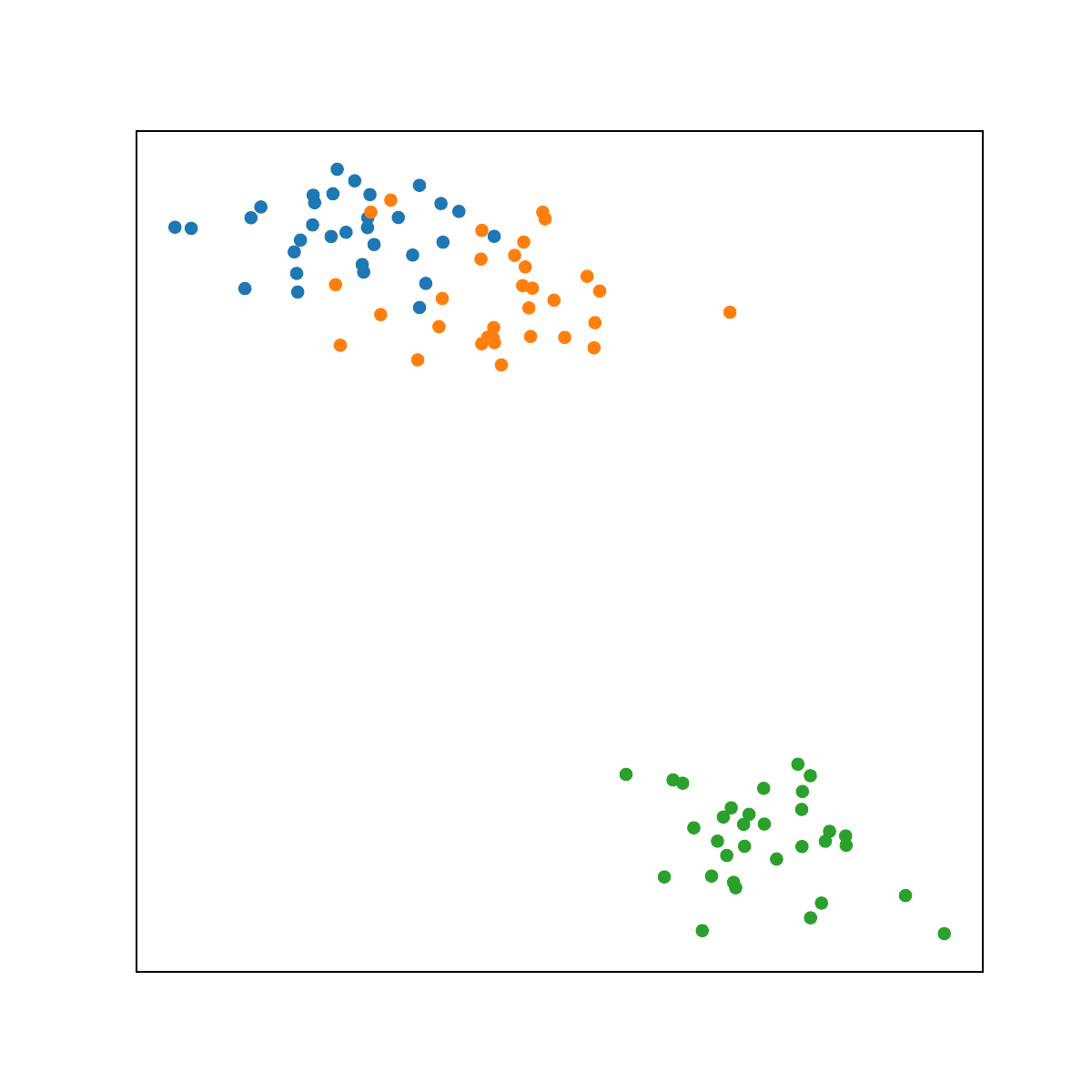}
  \caption{KDMCSE} 
\end{subfigure}
\begin{subfigure}[t]{0.45\linewidth} 
    \centering
  \includegraphics[width=\linewidth]{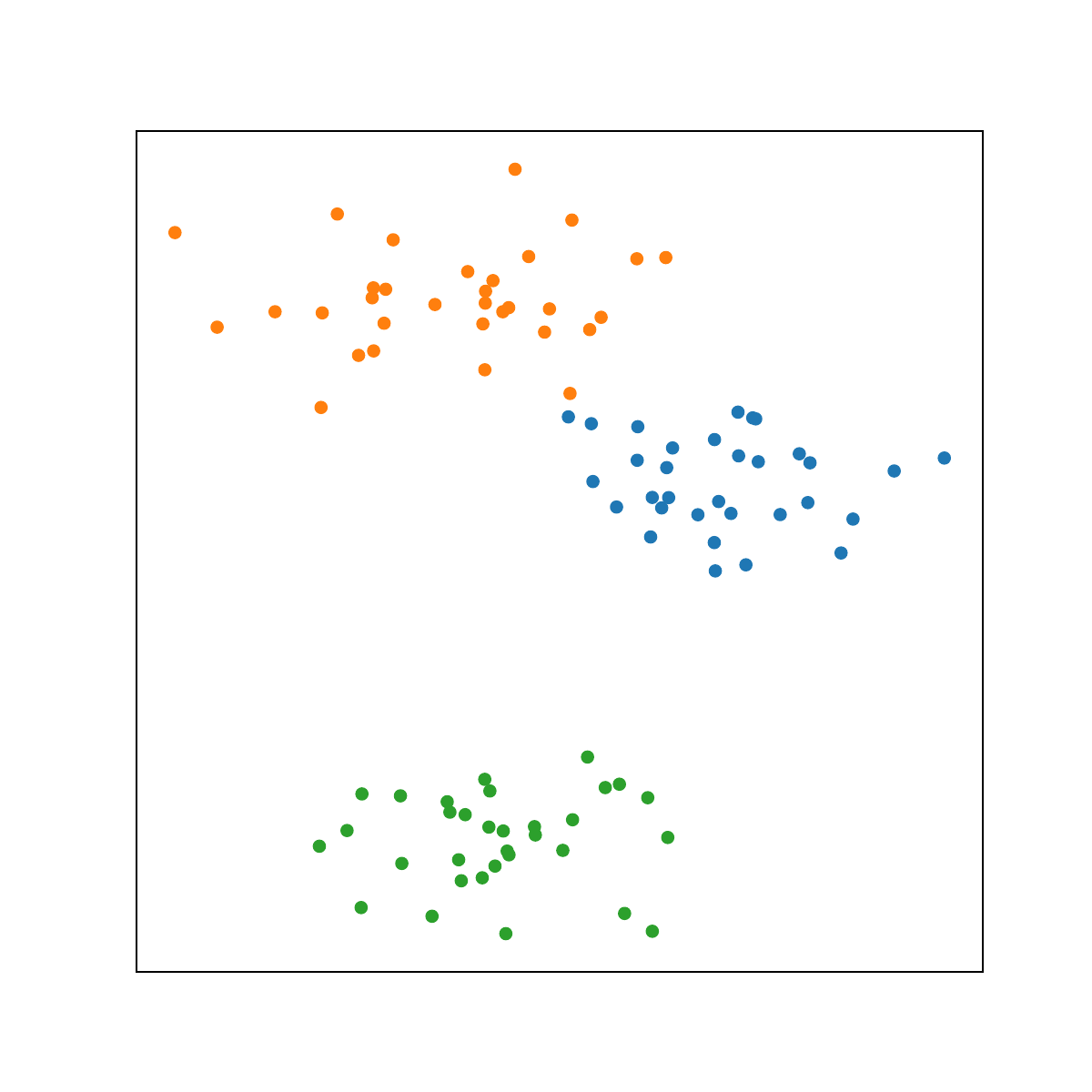}
  \caption{MCSE}
\end{subfigure}
\caption{\textcolor{orange}{Sentence 1}: A tattooed woman in a black dress holds a drink while sitting at a table in a dimly-lit room. \textcolor{blue}{Sentence 2}: Scantily clad woman in black waits in a restaurant with other patrons. \textcolor{green}{Sentence 3}: Woman bending down to pick up a tennis ball outside in front of a wall with graffiti on it.}
\label{fig:example5}
\end{figure}

In this appendix, we provide visualization of sentence embedding of our KDMCSE and MCSE models, illustrated in Figures~\ref{fig:example1},~\ref{fig:example2},~\ref{fig:example3},~\ref{fig:example4} and~\ref{fig:example5}. We visualize sentence representations by repeatedly processing three closely related sentences through the MCSE and KDMCSE models, employing various dropout masks to produce diverse representations for each sentence. Subsequently, we normalize the embeddings and apply t-SNE for dimensionality reduction, facilitating a clearer visualization of the data. For each example, we take three captions: two captions are almost similar in semantics (labeled with \textcolor{orange}{orange} and \textcolor{blue}{blue}), and the remaining caption (labeled with \textcolor{green}{green}) is different from the other two captions.

%% file: sections/table2.tex
\begin{table*}[ht]
 \begin{center}
 \scalebox{0.75}{
  \begin{tabular}{lccccccc|c}
    \toprule
    \textbf{Model} & \textbf{STS12} & \textbf{STS13} & \textbf{STS14} & \textbf{STS15} & \textbf{STS16} & \textbf{STS-B} & \textbf{SICK-R} & \textbf{Avg.$\uparrow$} \\
    \midrule
    \midrule

    KDMCSE-BERT &
     74.4 &
     83.1 &
     76.3 &
     83.7 &
     78.8 &
     81.3 &
     73.0 &
     78.6 \\ 
    KDMCSE without AdapACSE &
     73.8 &
     81.3 &
     75.4 &
     83.2 &
     79.0 &
     80.5 &
     74.1 &
     78.2 \\
    KDMCSE without Threshold Filtering &
     73.4 &
     82.7 &
     76.6 &
     83.9 &
     78.8 &
     80.8 &
     72.9 &
     78.4 \\
    \bottomrule
 \end{tabular}}\\
   \vspace{1mm}

  \caption{Ablation results on STS tasks. STS-B: STS Benchmark, SICK-R: SICK-Relatedness, Avg.: average across $7$ tasks. The models are trained with the BERT encoder and in \textit{wiki-flickr} setting.}
  \label{tab:re2}
  \end{center}
  \vspace{-2pt}
\end{table*}